\documentclass{article} 
\usepackage[accepted]{icml2022}
\usepackage{microtype}
\usepackage{graphicx}
\usepackage{subfigure}
\usepackage{caption} 
\usepackage{booktabs} 
\usepackage{placeins}
\usepackage{adjustbox}
\usepackage{enumitem}
\usepackage{multicol}
\usepackage{multirow}
\usepackage{color, colortbl}
\definecolor{MyLightGray}{gray}{0.925}
\definecolor{MyDarkGray}{gray}{0.55}
\usepackage{xcolor}         %
\definecolor{myblue}{rgb}{0.00, 0.45, 0.70}
\definecolor{mygreen}{rgb}{0.01, 0.62, 0.45}
\definecolor{myred}{rgb}{0.84, 0.37, 0.00}

\PassOptionsToPackage{hyphens}{url}\usepackage{hyperref}
\hypersetup{
    colorlinks = false,
    linkbordercolor = mygreen,
    citebordercolor = mygreen,
    urlbordercolor = mygreen,
}
\usepackage{url}
\usepackage{wrapfig}

\usepackage[symbol]{footmisc}
\renewcommand{\thefootnote}{\fnsymbol{footnote}}

\icmltitlerunning{Prioritized Training on Points that are Learnable, Worth Learning, and Not Yet Learnt}

\usepackage{mathtools} 
\usepackage{mathcommand}
\usepackage{amssymb}
\usepackage{amsfonts} 
\usepackage{nicefrac}

\DeclareMathOperator{\opExpectation}{\mathbb{E}}
\newcommand{\E}[2]{\opExpectation_{#1} \left [ \ifblank{#2}{\:\cdot\:}{#2} \right ]}
\newcommand{\simpleE}[1]{\opExpectation_{#1}} 

\providecommand\given{\vert}

\usepackage[noend]{algpseudocode}

\newcommand\MidSymbol[1][]{%
\nonscript\:#1
\allowbreak
\nonscript\:
\mathopen{}}

\DeclareMathOperator{\opInformationContent}{h}
\DeclarePairedDelimiterXPP{\ICof}[1]{\opInformationContent}{(}{)}{}{%
    \ifblank{#1}{\:\cdot\:}{#1}
}

\DeclareMathOperator{\opEntropy}{H}
\DeclareMathOperator{\opPEntropy}{h}
\DeclarePairedDelimiterXPP{\Hof}[1]{\opEntropy}{[}{]}{}{%
    \renewcommand\given{\MidSymbol[\delimsize\vert]}
    \ifblank{#1}{\:\cdot\:}{#1}
}
\DeclarePairedDelimiterXPP{\hof}[1]{\opPEntropy}{[}{]}{}{%
    \renewcommand\given{\MidSymbol[\delimsize\vert]}
    \ifblank{#1}{\:\cdot\:}{#1}
}
\DeclarePairedDelimiterXPP{\Lof}[1]{L}{[}{]}{}{%
    \renewcommand\given{\MidSymbol[\delimsize\vert]}
    \ifblank{#1}{\:\cdot\:}{#1}
}
\DeclarePairedDelimiterXPP{\htof}[1]{\opPEntropy_{\text{true}}}{[}{]}{}{%
    \renewcommand\given{\MidSymbol[\delimsize\vert]}
    \ifblank{#1}{\:\cdot\:}{#1}
}
\DeclarePairedDelimiterXPP{\xHof}[1]{\opEntropy}{(}{)}{}{%
    \ifblank{#1}{\:\cdot\:}{#1}
}

\DeclareMathOperator{\opMI}{I}
\DeclarePairedDelimiterXPP{\MIof}[1]{\opMI}{[}{]}{}{%
    \renewcommand\given{\MidSymbol[\delimsize\vert]}
    \ifblank{#1}{\:\cdot\:}{#1}
}
\DeclareMathOperator{\opPMI}{pmi}
\DeclarePairedDelimiterXPP{\PMIof}[1]{\opPMI}{[}{]}{}{%
    \renewcommand\given{\MidSymbol[\delimsize\vert]}
    \ifblank{#1}{\:\cdot\:}{#1}
}

\DeclarePairedDelimiterXPP{\CrossEntropy}[2]{\opEntropy}{(}{)}{}{%
    \ifblank{#1#2}{\:\cdot\: \MidSymbol[\delimsize\Vert] \:\cdot\:}{#1 \MidSymbol[\delimsize\Vert] #2}
}

\DeclareMathOperator{\opKale}{D_\mathrm{KL}}
\DeclarePairedDelimiterXPP{\Kale}[2]{\opKale}{(}{)}{}{%
    \ifblank{#1#2}{\:\cdot\: \MidSymbol[\delimsize\Vert] \:\cdot\:}{#1 \MidSymbol[\delimsize\Vert] #2}
}

\DeclareMathOperator{\opp}{p}
\DeclarePairedDelimiterXPP{\pof}[1]{\opp}{(}{)}{}{%
    \renewcommand\given{\MidSymbol[\delimsize\vert]}
    \ifblank{#1}{\:\cdot\:}{#1}
}
\DeclarePairedDelimiterXPP{\ptof}[1]{\opp_{\text{true}}}{(}{)}{}{%
    \renewcommand\given{\MidSymbol[\delimsize\vert]}
    \ifblank{#1}{\:\cdot\:}{#1}
}
\DeclarePairedDelimiterXPP{\pevalof}[1]{\opp_{\text{eval}}}{(}{)}{}{%
    \renewcommand\given{\MidSymbol[\delimsize\vert]}
    \ifblank{#1}{\:\cdot\:}{#1}
}

\DeclarePairedDelimiterXPP{\ppof}[1]{\opp'}{(}{)}{}{%
    \renewcommand\given{\MidSymbol[\delimsize\vert]}
    \ifblank{#1}{\:\cdot\:}{#1}
}

\DeclarePairedDelimiterXPP{\pcof}[2]{\opp_{#1}}{(}{)}{}{%
    \renewcommand\given{\MidSymbol[\delimsize\vert]}
    \ifblank{#2}{\:\cdot\:}{#2}
}

\DeclareMathOperator{\opq}{q}
\DeclarePairedDelimiterXPP{\qof}[1]{\opq}{(}{)}{}{%
    \renewcommand\given{\MidSymbol[\delimsize\vert]}
    \ifblank{#1}{\:\cdot\:}{#1}
}

\DeclarePairedDelimiterXPP{\varHof}[2]{\opEntropy_{\ifblank{#1}{\:\cdot\:}{#1}}}{[}{]}{}{%
    \renewcommand\given{\MidSymbol[\delimsize\vert]}
    \ifblank{#2}{\:\cdot\:}{#2}
}

\DeclarePairedDelimiterXPP{\xvarHof}[2]{\opEntropy_{\ifblank{#1}{\:\cdot\:}{#1}}}{(}{)}{}{%
    \renewcommand\given{\MidSymbol[\delimsize\vert]}
    \ifblank{#2}{\:\cdot\:}{#2}
}

\newcommand{\D}{\mathcal{D}}
\newcommand{\Dtrain}{\mathcal{D}_\text{t}}

\newcommand{\Dval}{\mathcal{D}_\text{ho}}

\newcommand{\xval}{\textbf{x}^\text{ho}}

\newcommand{\yval}{\textbf{y}^\text{ho}}

\newcommand{\xvali}{x^\text{ho}}

\newcommand{\yvali}{y^\text{ho}}

\newcommand\blfootnote[1]{%
  \begingroup
  \renewcommand\thefootnote{}\footnote{#1}%
  \addtocounter{footnote}{-1}%
  \endgroup
}

\usepackage{FiraMono}
\usepackage[T1]{fontenc}

\algdef{SE}[SUBALG]{Indent}{EndIndent}{}{\algorithmicend\ }%
\algtext*{Indent}
\algtext*{EndIndent}

\everypar{\looseness=-1}
\DeclareMathOperator*{\argmax}{arg\,max}
\DeclareMathOperator*{\argmin}{arg\,min}

\usepackage[many]{tcolorbox}    
\newtcolorbox{mybox}[1]{%
    tikznode boxed title,
    enhanced,
    arc=0mm,
    interior style={white},
    attach boxed title to top center= {yshift=-\tcboxedtitleheight/2},
    fonttitle=\bfseries,
    colbacktitle=white,coltitle=black,
    boxed title style={size=normal,colframe=white,boxrule=0pt},
    title={#1}}

\begin{document}
\twocolumn[

\icmltitle{Prioritized Training on Points that are Learnable, \\ Worth Learning, and Not Yet Learnt }

\icmlsetsymbol{equal}{*}

\begin{icmlauthorlist}
\icmlauthor{S\"{o}ren Mindermann}{equal,OATML}
\icmlauthor{Jan Brauner}{equal,OATML}
\icmlauthor{Muhammed Razzak}{equal,OATML}
\icmlauthor{Mrinank Sharma}{equal,Oxfordstats}
\icmlauthor{Andreas Kirsch}{OATML}
\icmlauthor{Winnie Xu}{Cohere,Toronto}
\icmlauthor{Benedikt H\"{o}ltgen}{OATML}
\icmlauthor{Aidan N. Gomez}{Cohere,OATML}
\icmlauthor{Adrien Morisot}{Cohere}
\icmlauthor{Sebastian Farquhar}{OATML}
\icmlauthor{Yarin Gal}{OATML}
\end{icmlauthorlist}

\icmlaffiliation{OATML}{OATML, Department of Computer Science, University of Oxford}
\icmlaffiliation{Oxfordstats}{Department of Statistics, University of Oxford}
\icmlaffiliation{Cohere}{Cohere}
\icmlaffiliation{Toronto}{University of Toronto, performed at Cohere}

\icmlcorrespondingauthor{S\"{o}ren Mindermann}{soeren.mindermann@gmail.com}

\vskip 0.2in
]

\begin{abstract}

Training on web-scale data can take months. But most computation and time is wasted on \mbox{redundant} and noisy points that are already learnt or not learnable. To accelerate training, we introduce \textit{Reducible Holdout Loss Selection} (RHO-LOSS), a simple but principled technique which selects approximately those points for training that most reduce the model's generalization loss. As a result, RHO-LOSS mitigates the weaknesses of existing data selection methods: techniques from the optimization literature typically select ``hard’’ (e.g. high loss) points, but such points are often noisy (not learnable) or less task-relevant. Conversely, curriculum learning prioritizes ``easy’’ points, but such points need not be trained on once learnt. In contrast, RHO-LOSS selects points that are learnable, worth learning, and not yet learnt. RHO-LOSS trains in far fewer steps than prior art, improves accuracy, and speeds up training on a wide range of datasets, hyperparameters, and architectures (MLPs, CNNs, and BERT). On the large web-scraped image dataset \mbox{Clothing-1M}, RHO-LOSS trains in 18x fewer steps and reaches 2\% higher final accuracy than uniform data shuffling. 
\blfootnote{\hspace{-6.3mm} Code: \url{https://github.com/OATML/RHO-Loss} \vspace{0.1cm}}
\printAffiliationsAndNotice{\icmlEqualContribution}

\end{abstract}

\section{Introduction}
\label{sec:intro}

State-of-the-art models such as GPT-3 \citep{brown2020language}, CLIP \citep{radford2021learning}, and ViT \citep{dosovitskiy2021image} achieve remarkable results by training on vast amounts of web-scraped data. 
But despite intense parallelization, training such a model takes weeks or months \citep{radford2021learning, chowdhery2022palm}.
Even practitioners who work with smaller models face slow development cycles, due to numerous iterations of algorithm design and hyperparameter selection. 
As a result, the total time required for training is a core constraint in the development of such deep learning models.

If it further sped up training, practitioners with sufficient resources would use much larger batches and distribute them across many more machines \citep{anil2018large}. However, this has rapidly diminishing returns \citep{lecun2012efficient}, to a point where adding machines does not reduce training time \citep{mccandlish2018empirical, anil2018large}---see e.g. GPT-3 and PaLM \citep{chowdhery2022palm}.

\begin{figure}
    \vspace{-1mm}
    \centering
    \hspace*{-1.5mm}\includegraphics[width=\columnwidth]{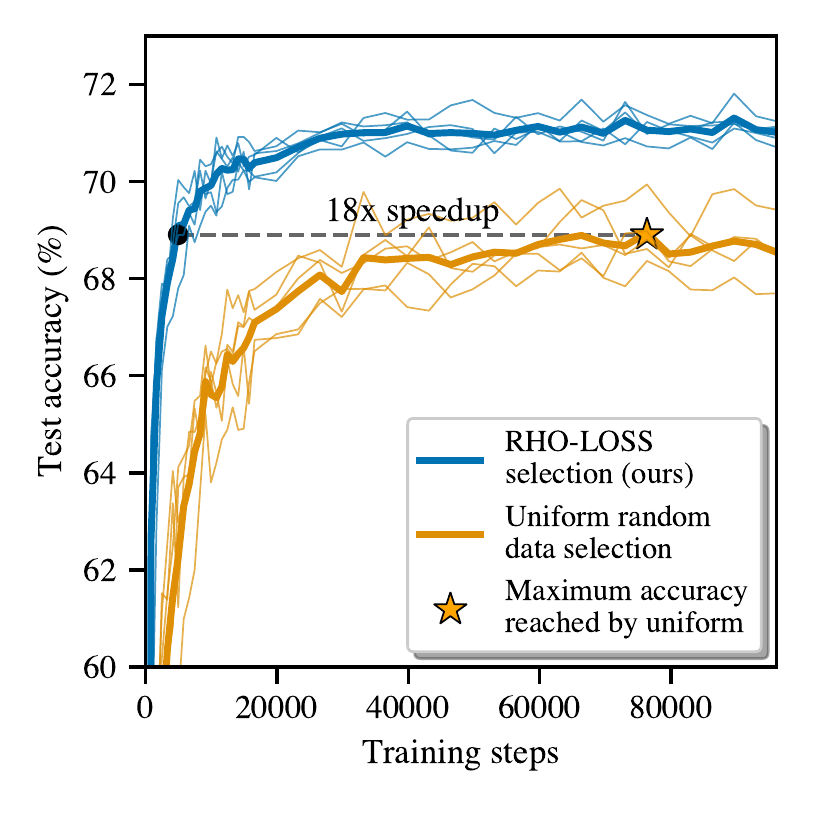}
    \vspace{-2mm}
    \caption{
    \textbf{Speedup on large-scale classification of \mbox{web-scraped} data (Clothing-1M)}. 
    RHO-LOSS trains all architectures with fewer gradient steps than standard uniform data selection (i.e. shuffling), helping reduce training time. Thin lines: ResNet-50, MobileNet v2, DenseNet121, Inception v3, GoogleNet, mean across seeds. Bold lines: mean across all architectures.}
    \label{fig:clothing-1M}
    \vspace{0mm}
\end{figure}

Additional machines can, however, still speed up training by filtering out less useful samples \citep{alain2015variance}.
Many web-scraped samples are \textcolor{myblue}{\textit{noisy}}, i.e. their label is incorrect or inherently ambiguous. For example, the text associated with a web-scraped image is rarely an accurate description of the image. 
Other samples are learnt quickly and are then \textcolor{myred}{\textit{redundant}}. Redundant samples are commonly part of object classes that are over-represented in web-scraped data \citep{tian2021divide} and they can often be left out without losing performance. 
Given that web-scraped data is plentiful---often enough to finish training in a single epoch \citep{komatsuzaki2019one, brown2020language}---one can afford to skip less useful points. 

However, there is no consensus on which datapoints are the most useful. Some works, including curriculum learning, suggest prioritizing \textit{easy} points with low \textcolor{myblue}{label noise} before training on all points equally \citep{bengio2009curriculum}. While 
this approach may improve convergence and generalization, it lacks a mechanism to skip points that are \textcolor{myred}{already learnt} (\textcolor{myred}{redundant}). Other works instead suggest training on points that are \textit{hard} for the model, thereby avoiding \textcolor{myred}{redundant} points, whose loss cannot be further reduced. Online batch selection methods \citep{loshchilov2015online, katharopoulos2018not, jiang2019accelerating, schaul2015prioritized} do so by selecting points with high loss or high gradient norm.

We show two failure modes of prioritising hard examples. Firstly, in real-world noisy datasets, high loss examples 
may be \textcolor{myblue}{mislabelled or ambiguous}.
Indeed, in controlled experiments, points selected by high loss or gradient norm are overwhelmingly those with \textcolor{myblue}{noise-corrupted} labels. Our results show that this failure mode degrades performance severely. More subtly, we show that some samples are hard because they are \textcolor{mygreen}{outliers}---points with unusual features that are \textcolor{mygreen}{less likely to appear at test time}. For the aim of reducing test loss, such points are \textcolor{mygreen}{less \textit{worth learning}}.

To overcome these limitations, we introduce \emph{reducible holdout loss selection} (RHO-LOSS). We propose a selection function grounded in probabilistic modelling that quantifies by how much each point would reduce the generalizaiton loss if we were to train on it, \textit{without actually training on it}. 
We show that optimal points for reducing holdout loss are \textcolor{myblue}{non-noisy}, \textcolor{myred}{non-redundant}, and \textcolor{mygreen}{task-relevant}. 
To approximate optimal selection, we derive an efficient and easy to implement selection function: the reducible holdout loss.

We explore RHO-LOSS in extensive experiments on 7 datasets. We evaluate the reduction in required training steps compared to uniform sampling and state-of-the-art batch selection methods. 
Our evaluation includes Clothing-1M, the main large benchmark with noisy, web-scraped labels, matching our main application.
RHO-LOSS reaches target accuracy in $18$x fewer steps than uniform selection and achieves $2\%$ higher final accuracy (Fig.~\ref{fig:clothing-1M}). 
Further, RHO-LOSS consistently outperforms prior art and speeds up training across datasets, modalities, architectures, and hyperparameter choices. 
Explaining this, we show that methods selecting ``hard'' points prioritize noisy and less relevant examples. In contrast, RHO-LOSS chooses \textcolor{myblue}{low-noise}, \textcolor{mygreen}{task-relevant}, \textcolor{myred}{non-redundant} points---points that are \textcolor{myblue}{learnable}, \textcolor{mygreen}{worth learning}, and \textcolor{myred}{not yet learnt}.

\section{Background: Online Batch Selection}
\label{sec:background}

Consider a model $\pof{y\given x;\theta}$ with parameters $\theta$ training on data $\mathcal{D} = \{(x_i, y_i)\}_{i=1}^n$ using stochastic gradient descent (SGD). At each training step $t$, we load a batch $b_t$ of size $n_b$ from $\mathcal{D}$. 
In online batch selection \citep{loshchilov2015online}, we uniformly pre-sample a larger batch $B_t$ of size $n_B>n_b$. Then, we construct a smaller batch $b_t$ that consists of the top-ranking $n_b$ points in $B_t$ ranked by a label-aware selection function $S(x_i, y_i)$. We perform a gradient step to minimize a mini-batch loss $L(y_i, \pof{y_i\given x_i;\theta})$ summed over $i\in b_t$. The next large batch $B_{t+1}$ is then pre-sampled from $\mathcal{D}$ without replacement of previously sampled points (i.e. random shuffling: replacement when the next epoch starts). 

\section{Reducible Holdout Loss Selection}
\label{sec:theory}

Previous online batch selection methods, such as loss or gradient norm selection, aim to select points that, if we were to train on them, would minimize the \textit{training set} loss. \citep{loshchilov2015online,katharopoulos2018not, kawaguchi2020ordered, alain2015variance}. Instead, we aim to select points that minimize the loss on a \textit{holdout set}. 
It would be too expensive to naively train on every candidate point and evaluate the holdout loss each time. 
In this section, we show how to (approximately) find the points that would most reduce the holdout loss if we were to train the current model on them, without actually training on them.

For simplicity, we first assume only one point \mbox{$(x,y)\in B_t$} is selected for training  at each time step $t$ (we discuss selection of multiple points below). $\pof{y' \given x'; \Dtrain}$ is the predictive distribution of the current model, where $\Dtrain$ is the sequence of data the model was trained on before training step $t$. $\Dval = \{(\xvali_i, \yvali_i)\}_{i=1}^{n^{\text{ho}}}$, written as $\xval$ and $\yval$ for brevity, is a holdout set drawn from the same data-generating distribution $\ptof{x',y'}$ as the training set $\mathcal{D}$. We aim to acquire the point $(x, y)\in B_t$ that, if we were to train on it, would minimize the negative log-likelihood/cross-entropy loss on the holdout set: 
\begin{flalign} \label{PSI1}
    \argmin_{(x,y) \in B_t} - \log \pof{\yval \given \xval; \Dtrain \cup (x, y)}.
\end{flalign}

For a model using a point estimate of $\theta$ (such as an MLE or MAP), rather than a distribution over $\theta$, 
the holdout loss factorises and (up to a constant factor) forms a Monte Carlo approximation of the expected loss under
$\text{p}_{\text{true}}$: $\simpleE{\ptof{x',y'}}[\Lof{y' \given x'; \Dtrain \cup (x,y)}] \approx \frac{1}{|\Dval|} \sum_{(\xvali_i,\yvali_i) \in \Dval} \Lof{\yvali_i \given \xvali_i; \Dtrain \cup (x, y)}$, where $\Lof{\cdot}$ denotes the cross-entropy loss: $\Lof{y \given x}:=- \log \pof{ y \given x}$.

\paragraph{Deriving a tractable selection function.}
We now derive a tractable expression for the term in Eq.~\eqref{PSI1} that does not require us to train on each candidate point $(x, y)\in B_t$ and then evaluate the loss on $\Dval$. To make our claims precise and our assumptions transparent, we use the language of Bayesian probability theory. We treat model parameters as a random variable with prior $\pof{\theta}$ and infer a posterior $\pof{\theta|\Dtrain}$ using the already-seen training data $\Dtrain$. The model has a predictive distribution $\pof{y|x,\Dtrain} =~\int_\theta
\pof{y|x,\theta}\pof{\theta|\Dtrain}d\theta$. When using a point estimate of $\theta$, the predictive distribution can be written as an integral with respect to a Dirac delta.

Using Bayes rule and the conditional independence \mbox{$\pof{y_i \given x_i, x_j; \Dtrain} = \pof{y_i \given x_i; \Dtrain}$}, we can derive a tractable selection function from Eq.~(\ref{PSI1}). For readability, we switch the sign of the selection function, later changing the minimization to a maximization.
\begin{align} 
    &
    \log \pof{\yval \given \xval; \Dtrain \cup (x, y)}\notag\\[0.5ex]
    & = \log \frac{ \pof{y \given x; \xval, \yval, \Dtrain} \pof{\yval \given \xval, x; \Dtrain} }{ \pof{y \given x, \xval; \Dtrain}}  
    \text{\begin{tabular}{c}
        Bayes rule  
    \end{tabular}}\notag
    \\[0.5ex]
    & = \log  \frac{ \pof{y \given x; \yval, \xval, \Dtrain} \pof{\yval \given \xval; \Dtrain} }{ \pof{y \given x; \Dtrain} } 
    \text{\begin{tabular}{c}
        conditional\\
        independence
    \end{tabular}}\notag \\[0.5ex]
    & \propto \Lof{ y \given x; \Dtrain} - \Lof{y \given x; \Dval, \Dtrain}, \label{RHOLOSS_wo_approximation}
\end{align}
where in the final line, we dropped terms independent of $(x,y)$, rearranged, and applied the definition of $L[\cdot]$.

As exact Bayesian inference (conditioning on $\Dtrain$ or $\Dval$) is intractable in neural networks \citep{pmlr-v37-blundell15}, we fit the models with SGD instead (\textbf{Approximation 1}). We study the impact of this approximation in Section~\ref{sec:impact_of_approximations}. The first term, $\Lof{ y \given x; \Dtrain}$, is then the \textit{training loss} on the point $(x,y)$ using the current model trained on $\Dtrain$. The second term, $\Lof{y \given x; \Dval, \Dtrain}$, is the loss of a model trained on $\Dtrain$ and $\Dval$.

Although the selection function in Eq.~\eqref{RHOLOSS_wo_approximation} is tractable, it is still somewhat expensive to compute, as both terms must be updated after each acquisition of a new point. However, we can approximate the second term with a model trained only on the holdout dataset, \mbox{$\Lof{y \given x; \Dval, \Dtrain} \approx \Lof{y \given x; \Dval}$} (\textbf{Approximation 2}). This approximation saves a lot of compute: it is now sufficient to compute the term once before the first epoch of training. Later on, we show that this approximation empirically does not hurt performance on any tested dataset and even has some desired properties (Section~\ref{sec:impact_of_approximations} and Appendix~\ref{app:update_irreducible_2}). We term $\Lof{y \given x; \Dval}$ the \textit{irreducible holdout loss} (IL) since it is the remaining loss on point $(x,y) \in  \mathcal{D}$ after training on the holdout set $\Dval$; in the limit of $\Dval$ being large, it would be the lowest loss that the model can achieve without training on $(x,y)$. Accordingly, we name our approximation of Eq.~\eqref{RHOLOSS_wo_approximation} the \textit{reducible holdout loss}---the difference between the training loss and the irreducible holdout loss (IL).

Our method still requires us to train a model on a holdout set, but a final approximation greatly reduces that cost. We can efficiently compute the IL with an ``irreducible loss model" (IL model) that is smaller than the target model and has low accuracy (\textbf{Approximation 3}). We show 
this and explain it in Sections~\ref{sec:impact_of_approximations}, \ref{sec:cheap_irlo}, and \ref{sec:properties}. Counterintuitively, the reducible holdout loss can therefore be negative. Additionally, one IL model can be reused for many target model runs, amortizing its cost (Section~\ref{sec:cheap_irlo}). For example, we trained all 40 seeds of 5 target architectures in Fig.~\ref{fig:clothing-1M} using a single ResNet18 IL model. Further, this model trained for 37x fewer steps than each target model (reaching only 62\% accuracy). Section~\ref{sec:related} details further possible efficiency improvements. 

In summary, selecting a point that minimizes the holdout loss in Eq.~\eqref{PSI1}, for a model trained on $\Dtrain$, can be approximated with the following easy-to-compute objective:

\begin{mybox}{Reducible holdout loss selection (RHO-LOSS)}
\begin{flalign} \label{eq:rholoss}
    \hspace{-2.5mm} \argmax_{(x,y) \in B_t} \:\:\: \overbrace{\underbrace{\Lof{y \given x; \Dtrain}}_{\text{training loss}}  ~~~-\underbrace{\Lof{y \given x; \Dval}}_{\text{irreducible holdout loss (IL)}}}^{\text{reducible holdout loss}}
\end{flalign}
\end{mybox}

Although we required additional data $\Dval$, this is not essential for large (Section~\ref{sec:experiments}.0) nor small (Section~\ref{sec:cheap_irlo}) datasets.

\begin{algorithm}[t]
	\caption{Reducible holdout loss selection (RHO-LOSS)}
    \label{alg:RHOLOSS}
	\newcommand{\StatexIndent}[1][3]{%
  \setlength\@tempdima{\algorithmicindent}%
  \Statex\hskip\dimexpr#1\@tempdima\relax}
	\begin{algorithmic}[1] %
	     \State \textbf{Input:} 
	      Small model $\pof{y\given x; \Dval}$ trained on a holdout set $\Dval$, batch size $n_b$, large batch size $n_B>n_b$, \mbox{learning rate $\eta$}.
	    \For{$(x_i, y_i)$ in \texttt{training set}
	    }
	    \State \texttt{IrreducibleLoss[i]} $\gets \Lof{y_i\given x_i; \Dval} $
	    \EndFor
	    \item[] Initialize parameters $\theta^0$ and $t=0$
	    \For{$t=0, 1, \ldots$} 
	    \State Randomly select a large batch $B_t$ of size $n_B$.
	    \State \parbox[t]{0.9\linewidth}{$\forall i\in B_t$, compute \texttt{Loss[i]}, the train loss of point $i$ given parameters $\theta^t$} 
	    \vspace{0.45em}
	    \State \parbox[t]{0.9\linewidth}{$\forall i\in B_t$, compute \texttt{RHOLOSS[i]}$\gets$ { $\texttt{Loss[i]}-\texttt{IrreducibleLoss[i]}$}} \vspace{0.45em}
	    \State \parbox[t]{0.9\linewidth}{$b_t\gets$ top-$n_b$ samples in $B_t$ in terms of 
	   {$\texttt{RHOLOSS}$}.}\vspace{0.45em}
	   \State $g_t\gets$ mini-batch gradient on $b_t$ using parameters $\theta^t$
	   \State $\theta^{t+1} \gets \theta^t -\eta g_t$
	   \EndFor

	\end{algorithmic}
	\vspace{-1mm}
\end{algorithm}

\paragraph{Understanding reducible loss.} We now provide intuition on why reducible holdout loss selection (RHO-LOSS) avoids 
\textcolor{myred}{redundant}, \textcolor{myblue}{noisy}, and \textcolor{mygreen}{less relevant} points.
\mbox{\textbf{i) Redundant points.}} We call a training point redundant when the model has already learnt it, i.e. its training loss cannot be further reduced. Since \textcolor{myred}{redundant} points have \textcolor{myred}{low training loss}, and the reducible loss is always less than the training loss (Eq.~(\ref{eq:rholoss})), such points have low reducible loss and are not selected. 
And if the model forgets them, they are revisited in the next epoch. 
\textbf{ii) Noisy points.} While 
prior methods select based on high training loss (or gradient norm), not all points with high loss are informative---some may have an \textcolor{myblue}{ambiguous or incorrect} (i.e. \textcolor{myblue}{noisy}) label. The labels of such points cannot be predicted using the holdout set~\citep{chen2019understanding}. Such points have \textcolor{myblue}{high IL} and, consequently, low reducible loss. These \textcolor{myblue}{noisy} points are less likely to be selected compared to equivalent points with less noise. \textbf{iii) Less relevant points.} Loss-based selection has an additional pitfall. The training loss is likely higher for \textcolor{mygreen}{outliers} in input space---values of $x$ far from most of the training data, in regions with \textcolor{mygreen}{low input density} under $\ptof{x}$.
Points with low $\ptof{x}$ should not be prioritized, all else equal. Consider an `outlier' $(x, y)$ and a non-outlier $(x', y')$, with $\ptof{x}< \ptof{x'}$ but \textit{equal} training loss $\Lof{y|x;\Dtrain}=\Lof{y'|x';\Dtrain}$. As the holdout set $\Dval$ is also drawn from $\text{p}_{\text{true}}$, $\Dval$ will contain fewer points from the region around $x$ in input space compared to the region around $x'$. Thus,  training on $(x, y)$ is likely to reduce the holdout loss (Eq.~(\ref{PSI1})) less, and so we prefer to train on the non-outlier $(x', y')$. In the specific sense described, $(x, y)$ is thus \textcolor{mygreen}{less \textit{relevant}} to the holdout set. As desired, RHO-LOSS deprioritizes $(x, y)$: since $\Dval$ contains few points from the region around $x$, the IL of $(x, y)$ will be large.

In short, RHO-LOSS \textit{deprioritizes} points that are \textcolor{myred}{redundant} (low training loss), \textcolor{myblue}{noisy} (high IL), or \textcolor{mygreen}{less relevant} to the holdout set (high IL). That is, RHO-LOSS \textit{prioritizes} points that are \textcolor{myred}{not yet learnt}, \textcolor{myblue}{learnable}, and \textcolor{mygreen}{worth learning}. We provide empirical evidence for these claims in Section~\ref{sec:properties}. See Algorithm~\ref{alg:RHOLOSS} for the implementation of RHO-LOSS.

\paragraph{Selecting multiple points concurrently.}
We showed which point is optimal when selecting a single point $(x,y)$. 
When selecting an entire batch $b_t$, we select the points with the top-$n_b$ scores from the randomly pre-sampled set $B_t$. 
This is nearly optimal when assuming that 
each point has little effect on the score of other points, which is often used as a simplifying assumption in active learning \citep{kirsch2019batchbald}. This assumption is much more reasonable in our case than in active learning because model predictions are not changed much by a single gradient step on one mini-batch.

\paragraph{Simple parallelized selection.}

For large-scale neural network training, practitioners with sufficient resources would use many more machines if it further sped up training \citep{anil2018large}. However, as more workers are added in synchronous or asynchronous gradient descent, the returns diminish to a point where adding more workers does not further improve wall clock time \citep{anil2018large, mccandlish2018empirical}. For example, there are rapidly diminishing returns for using larger batch sizes or distributing a given batch across more workers, for multiple reasons \citep{mccandlish2018empirical, keskar2016large}. The same holds for distributing the model across more workers along its width or depth dimension \citep{rasley2020deepspeed, shoeybi2019megatron, huang2019gpipe}. However, we can circumvent these diminishing returns by adding a new dimension of parallelization, namely, for data selection.

Since parallel \textit{forward} passes do not suffer from such diminishing returns, one can use extra workers to evaluate training losses in parallel \citep{alain2015variance}. The theoretical runtime speedup can be understood as follows. The cost per training step of computing the selection function on $B_t$ is $\frac{n_B}{3n_b}$ times as much as the cost of the forward-backward pass needed to train on $b_t$ since a forward pass requires at least $3$x less computation than a forward-backward pass \citep{jouppi2017datacenter}. One can reduce the time for the selection phase almost arbitrarily by adding more workers that compute training losses using a copy of the model being trained. The limit is reached when the time for selection is dominated by the communication of parameter updates to workers. More sophisticated parallelization strategies allow reducing the time overhead even further (Section~\ref{sec:related}). To avoid assumptions about the particular strategy used, we report experiment results in terms of the required number of training epochs.

\vspace{-2mm}
\section{Experiments}\label{sec:experiments}

We evaluate our selection method on several datasets (both in controlled environments and real-world conditions) and show significant speedups compared to prior art, in the process shedding light on the properties of different selection functions. 

Recall that our setting assumes training time is a bottleneck but data is abundant---more than we can train on (see \citet{bottou2004large}). This is common e.g. for web-scraped data where state-of-the-art performance is often reached in less than half of one epoch \citep{komatsuzaki2019one, brown2020language}. As data is abundant, we can set aside a holdout set for training the IL model with little to no downside. For the large Clothing-1M dataset, we implement RHO-LOSS by training the IL model on 10\% of the training data, while all baselines are trained on the full 100\% of the training data. 
For the smaller datasets, we simulate abundance of data by reserving a holdout set and training \textit{all} methods only on the remaining data. 
However, RHO-LOSS also works on small datasets without additional data by double-using the training set (Section~\ref{sec:cheap_irlo}).

\paragraph{Datasets.}
We evaluate on 7 datasets: 
1) QMNIST \citep{qmnist-2019} extends MNIST \citep{lecun-98} with 50k extra images which we use as the holdout set.
2) On CIFAR-10 \citep{cifar10} we train on half of the training set and use the other half as a holdout to train the irreducible loss (IL) model. 3) CIFAR-100: same as CIFAR-10. 4) CINIC-10 \citep{darlow2018cinic} has 4.5x more images than CIFAR-100 and includes a holdout set and a test set with 90k images each. 5) Clothing-1M \citep{xiao2015learning}, which contains over 1 million 256x256-resolution clothing images from 14 classes. The dataset is fully web-scraped---a key application area of our work---and is the most widely accepted benchmark for image recognition with noisy labels \citep{algan2021image}. 
We use the whole training set for training and reuse 10\% of it to train the IL model. We further evaluate on two NLP datasets from GLUE \citep{wang2018glue}: 6) CoLA (grammatical acceptability) and 7) SST-2 (sentiment). We split their training sets as for CIFAR.

\begin{figure*}
    \centering
    \hspace*{-1.5mm}\includegraphics[width=0.95\linewidth]{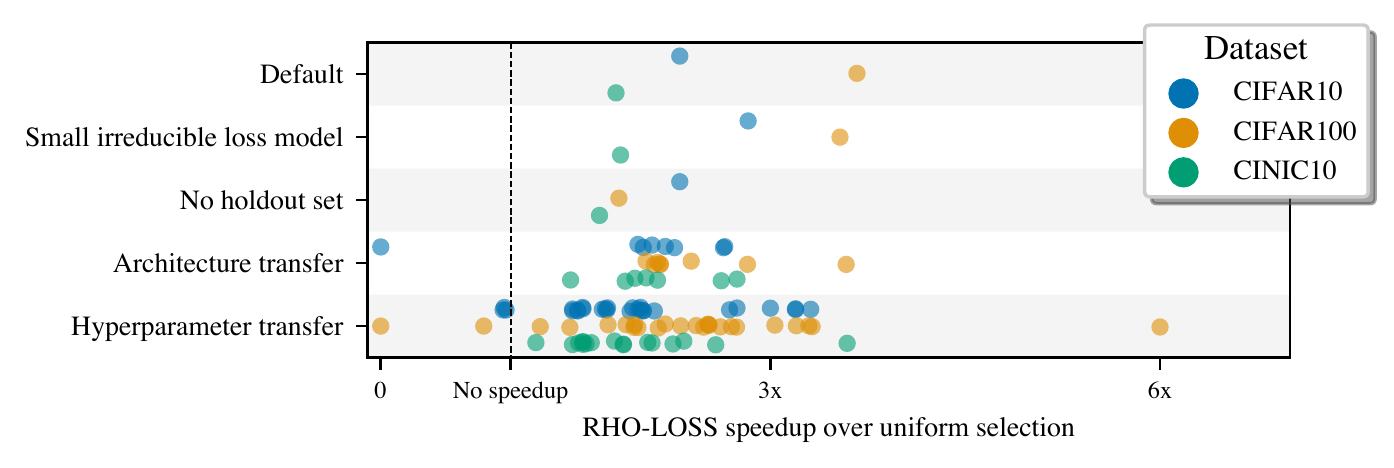}
    \caption{\textbf{The irreducible loss model can be small, trained with no holdout data, and reused across target architectures and hyperparameters.} Here, we use clean datasets, where speedups are smallest. The x-axis shows speedup, i.e. after how many fewer epochs RHO-LOSS exceeds the highest accuracy uniform selection achieves within 100 epochs. 
    Row 1 uses a ResNet18 as irreducible loss model. All other rows instead use a small, cheap CNN.
    Each dot shows an experiment with a given combination of irreducible loss model and target model (mean across 2-3 seeds for all but the last row).}
    \label{fig5}
\end{figure*}

\paragraph{Baselines.}
Aside from uniform sampling (without replacement, i.e. random shuffling), we also compare to selection functions that have achieved competitive performance in online batch selection recently: the (training) loss, as implemented by \citet{kawaguchi2020ordered}, gradient norm, and gradient norm with importance sampling (called \textit{gradient norm IS} in our figures), as implemented by \citet{katharopoulos2018not}.
We also compare to the core-set method Selection-via-Proxy (SVP) that selects data offline before training \citep{coleman2019selection}. We report results using maximum entropy SVP and select with the best-performing model, ResNet18. 
We further compare to four baselines from active learning, shown in Appendix~\ref{app:active_learning} as they assume labels are unobserved. Finally, we include selection using the negative IL (see Eq.~\ref{eq:rholoss}) to test if it is sufficient to only skip noisy and less relevant but not redundant points. %

\paragraph{Models and hyperparameters.}
To show our method needs no tuning, we use the PyTorch default hyperparameters (with the AdamW optimizer \citep{loshchilov2017decoupled}) and $\frac{n_b}{n_B}=0.1$. 
We test many additional hyperparameter settings in Figs.~\ref{fig5} (row 5) and \ref{fig:percent_train}. We test various architectures in Figs.~\ref{fig:clothing-1M} and \ref{fig5} (row 4). 
In all other figures, we use a 3 layer MLP for experiments on QMNIST, a ResNet-18 adapted for small images for CIFAR-10/CIFAR-100/CINIC-10, and a ResNet-50 for Clothing-1M. All models for Clothing-1M are pre-trained on ImageNet (standard for this dataset \cite{algan2021image}) and the IL model is \textit{always} a ResNet-18. For the NLP datasets, we use a pretrained ALBERT v2 \citep{lan2019albert}.
We always use the IL model checkpoint with lowest validation loss (not highest accuracy); this performs best. Details in Appendix~\ref{app:exp_details}.

\paragraph{Evaluation.} We measure speedup in terms of the number of epochs needed to reach a given test accuracy. We measure epochs needed, rather than wall clock time, as our focus is on evaluating a new selection function, not an entire training pipeline. 
Wall clock time depends primarily on the hardware used and implementation details that are beyond our scope. 
Most importantly, data selection is amenable to parallelization beyond standard data parallelism as discussed in Section~\ref{sec:theory}.

\subsection{Impact of Approximations}\label{sec:impact_of_approximations}


In Section~\ref{sec:theory}, we introduced a function for selecting exactly the points that most reduce the model's loss on a holdout set. To make this selection function efficient for deep neural networks, we made several approximations. Here, we study how these approximations affect the points selected, by successively introducing one approximation after the other.

Because the exact selection function (Eq.~\eqref{RHOLOSS_wo_approximation}) is intractable, we start with a close (and expensive) approximation as the gold standard (Approximation 0). To make Approximation 0 feasible, the experiments are conducted on an easy dataset---QMNIST (with 10\% uniform label noise and data duplication to mimic the properties of web-scraped data). We then successively introduce the Approximations 1, 2, and 3 described in Section~\ref{sec:theory}. To assess the impact of each approximation, we train a model without and with the approximations, 
and then compute the rank correlation (Spearman's correlation coefficient) of the selection function evaluated on each batch $B_t$. Across the first epoch, we present the mean of the rank correlations.
Since each approximation selects different data, the corresponding models become more different over time; this divergence causes some of the observed difference in the points they select.
See Appendix~\ref{sec:approximation_appendix} for details.


\begin{table}
\caption{
Spearman's rank correlation of rankings of data points by selection functions that are increasingly less faithful approximations of Eq.~\eqref{RHOLOSS_wo_approximation}, compared to the most faithful approximation. Approximations added from left to right. Mean across 3 seeds.}
\begin{adjustbox}{width=\columnwidth}
\begin{tabular}{@{}c|ccccl@{}}
\toprule
                  & Non-     & Not       & Not updating & Small  \\
                  & Bayesian & converged & IL model    & IL model \\ \midrule
Rank correlation   & 0.75 & 0.76 & 0.63 & 0.51 \\
\bottomrule
\end{tabular}
\label{table:overlap}
\end{adjustbox}
\end{table}

\textit{Approximation 0.} To get as close as possible to the Bayesian inference/conditioning used in Eq.~\eqref{RHOLOSS_wo_approximation}, we use a deep ensemble of 5 neural networks and train them \textit{to convergence} after every time step $t$ on the acquired dataset $b_t \cup \D_t$ \cite{wilson2020bayesian}.

\textit{Approximation 1: SGD instead of Bayesian inference/conditioning.} Approximation 0 is a close approximation of Eq.~\eqref{RHOLOSS_wo_approximation}, but training an ensemble to convergence at every step $t$ is far too expensive in practice. Starting from this gold-standard, we introduce two stronger approximations (1a and 1b) to move to standard neural network fitting with AdamW. 1a) First, we replace the ensemble with a single model, while still training to convergence at each time step. The Spearman's coefficient between this approximation and Approximation~0 is 0.75, suggesting similar points are selected (``Non-Bayesian'' in Table~\ref{table:overlap}).
1b) Next, we only take one gradient step on each new batch $b_t$. The Spearman's coefficient, when comparing this to Approximation 0, is 0.76 (``Not Converged" in Table \ref{table:overlap}).

\textit{Approximation 2. Not updating the IL model on the acquired data $\Dtrain$.} Second, we save compute by approximating $\Lof{y \given x; \Dtrain, \Dval}$ with $\Lof{y \given x; \Dval}$. The points selected are still similar to Approximation~0 (Spearman's coefficient 0.63, ``Not updating IL model'' in Table~\ref{table:overlap}). This approximation also performs well on other datasets (Appendix \ref{app:update_irreducible_2}).

\textit{Approximation 3: Small IL model.} Lastly, we use a model with 256 hidden units instead of 512 (4x fewer parameters) as the IL model and see again that similar points are selected (Spearman's coefficient 0.51 ). We study cheaper IL models in other forms and datasets in the next section.

\begin{figure*}
    \centering
    \includegraphics[width=\textwidth]{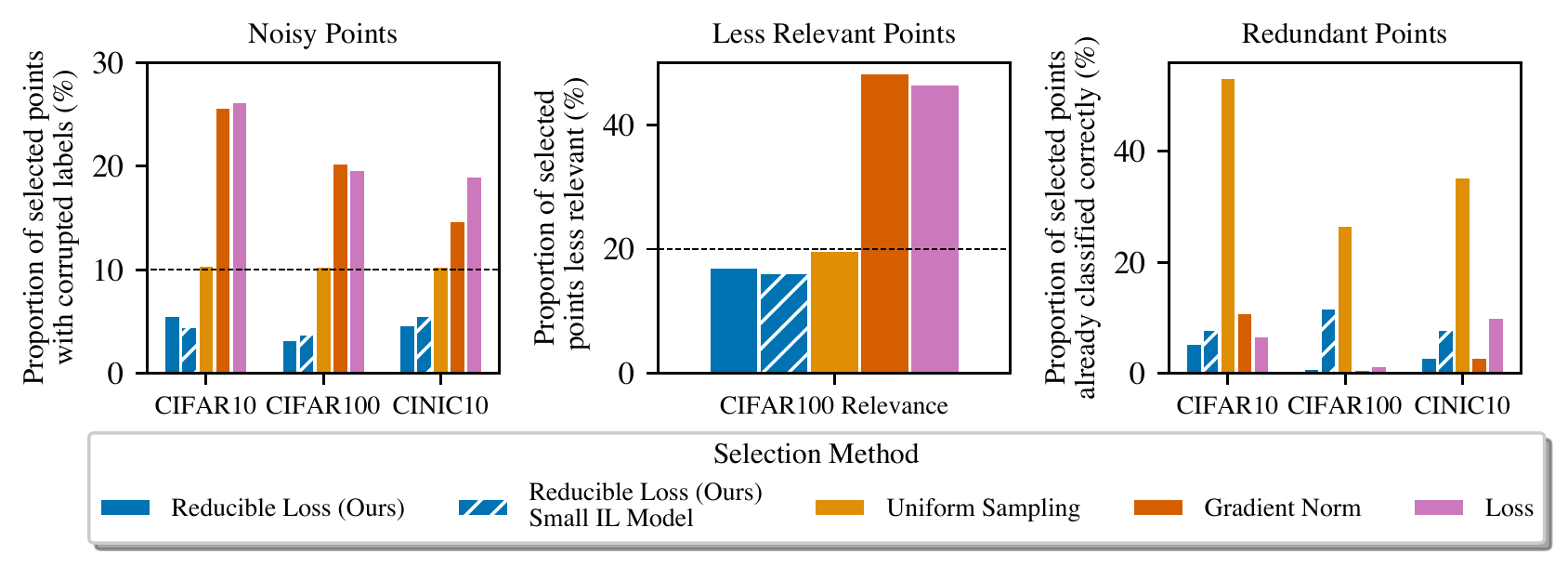}
    \caption{Properties of RHO-LOSS and other methods. 
    RHO-LOSS prioritizes points that are \textcolor{myblue}{non-noisy}, \textcolor{mygreen}{task-relevant}, and \textcolor{myred}{non-redundant}---even when the irreducible loss (IL) model is a small CNN. In contrast, loss and gradient norm prioritize noisy and less relevant points (while also avoiding redundant points). 
    \textbf{Left.} Proportion of selected points with corrupted labels. We added $10\%$ uniform label noise, i.e., we randomly switched each point's label with $10\%$ probability.
    \textbf{Middle.} Proportion of selected points from low relevance classes on CIFAR100 Relevance dataset.
    \textbf{Right.} Proportion of selected points that are already classified correctly, which is a proxy for redundancy.
    Mean over 150 epochs of training and 2-3 seeds. 
    }
    \label{fig:properties}
\end{figure*}

\subsection{Cheap Irreducible Loss Models \& Robustness}
\label{sec:cheap_irlo}

RHO-LOSS requires training an IL model on a holdout set, which poses additional costs. Here, we show how to minimize these costs and amortize them across many training runs of target models. The same experiments also  show the robustness of RHO-LOSS across architectures and hyperparameter settings. To fit our computational budget, we perform these experiments on moderate-sized clean benchmark datasets although RHO-LOSS produces greater speedups on noisy or redundant web-scraped data (see Section~\ref{sec:speedups}).

\paragraph{Irreducible loss models can be small and cheap.}

In our default setting (Fig.~\ref{fig5}, row 1), both the target model and IL model have the same architecture (ResNet-18). In rows 2 and below, we instead used a small CNN similar to LeNet as the IL model \citep{lecun1989backpropagation}. It has 21x fewer parameters and requires 29x fewer FLOP per forward pass than the ResNet-18. \textbf{The smaller IL model accelerates training as much or more than the larger model}, even though its final accuracy is far lower than the target ResNet-18 (11.5\% lower on CIFAR-10, 7\% on CIFAR-100, and 8.1\% on CINIC-10). We examine in Section~\ref{sec:properties} why this useful result holds.

\paragraph{Irreducible loss models without holdout data.}
Web-scraped datasets are often so large that even a small fraction of the overall data can be sufficient to train the IL model. E.g., in our experiments on Clothing-1M (Fig.~\ref{fig:clothing-1M}), the holdout set is only 10\% as large as the main train set. Additionally, we can train the IL model without any holdout data (Fig.~\ref{fig5}, row 3). We split the training set $\D$ into two halves and train an IL model on each half (still using small IL models). Each model computes the IL for the half of $\D$ that it was not trained on. Training two IL models costs no additional compute since each model is trained on half as much data compared to the default settings. 

\paragraph{Irreducible loss models can be reused to train different target architectures.}
We find that a single small CNN IL model accelerates the training of 7 target architectures (Fig.~\ref{fig5}, row 4): VGG11 (with batchnorm), GoogleNet, Resnet34, Resnet50, Densenet121, MobileNet-v2, Inception-v3. RHO-LOSS does not accelerate training on CIFAR-10 for VGG11, which is also the architecture on which uniform training performs the worst; i.e. RHO-LOSS empirically does not “miss” a good architecture. Not only is RHO-LOSS robust to architectures choice, a single IL model can also be \textit{reused by many practitioners} who use different architectures (as we did in Fig.~\ref{fig:clothing-1M}).

\paragraph{Irreducible loss models can be reused to train many targets in a hyperparameter sweep.}
We find that a single small CNN accelerates the training of ResNet-18 target models across a hyperparameter grid search (Fig.~\ref{fig5}, last row). We vary the batch size (160, 320, 960), learning rate (0.0001, 0.001, 0.01), and weight decay coefficient (0.001, 0.01, 0.1). RHO-LOSS speeds up training compared to uniform on nearly all target hyperparameters. The few settings in which it doesn't speed up training are also settings in which uniform training performs very poorly ($<30\%$ accuracy on CIFAR-100, $<80\%$ on CIFAR-10). 

\begin{table*}[ht]
\caption{Epochs required to reach a given target test accuracy (final accuracy in parentheses).  Figs.~\ref{fig:walltime}~and~~\ref{fig:walltime_nlp} (Appendix) show all training curves.
Some datasets have 10\% uniform label noise added. 
Results averaged across 2-4 seeds. Best performance in \textbf{bold}.  RHO-LOSS performs best in both epochs required and final accuracy. {\small \emph{NR}} indicates that the target accuracy was not reached. $^*$On CIFAR10/100, CoLA, and SST-2, only half of the data is used for training (Section~\ref{sec:experiments}.0).}
\begin{adjustbox}{width=\textwidth}
\begin{tabular}{c|c|ccccc|c|c}
\hline
\multirow{2}{*}{Dataset} & \multirow{2}{*}{Target Acc} & \multicolumn{7}{c}{Number of epochs method needs to reach target accuracy $\downarrow$ (Final accuracy in parentheses)} \\
 &  & Train Loss & Grad Norm & Grad Norm IS & SVP & Irred Loss & Uniform & RHO-LOSS \\ \hline

 Clothing-1M & 60.0\% & {\color{MyDarkGray}8} & {\color{MyDarkGray}13} & {\color{MyDarkGray}2} & {\small \color{MyDarkGray} \emph{NR}} & {\small \color{MyDarkGray} \emph{NR}} & 2 & \textbf{1} \\
  & 69.0\% & {\small \color{MyDarkGray} \emph{NR}} {\color{MyDarkGray} (65\%)} & {\small \color{MyDarkGray} \emph{NR}} {\color{MyDarkGray} (64\%)} & {\color{MyDarkGray}9} {\color{MyDarkGray} (70\%)} & {\small \color{MyDarkGray} \emph{NR}} {\color{MyDarkGray} (55\%)} & {\small \color{MyDarkGray} \emph{NR}} {\color{MyDarkGray} (48\%)} & 30 (70\%) & \textbf{2 (72\%)} \\

  \rowcolor{MyLightGray} CIFAR10$^*$ & 80.0\% & {\color{MyDarkGray}81} & {\small \color{MyDarkGray} \emph{NR}} & {\color{MyDarkGray}57} & {\small \color{MyDarkGray} \emph{NR}} & {\small \color{MyDarkGray} \emph{NR}} & 79 & \textbf{39} \\
  \rowcolor{MyLightGray} & 87.5\% & {\color{MyDarkGray}129} {\color{MyDarkGray} (90\%)} & {\small \color{MyDarkGray} \emph{NR}} {\color{MyDarkGray} (61\%)} & {\color{MyDarkGray}139} {\color{MyDarkGray} (89\%)} & {\small \color{MyDarkGray} \emph{NR}} {\color{MyDarkGray} (55\%)} & {\small \color{MyDarkGray} \emph{NR}} {\color{MyDarkGray} (60\%)} & {\small \emph{NR}} (87\%) & \textbf{65 (91\%)} \\

  CIFAR10$^*$ & 75.0\% & {\small \color{MyDarkGray} \emph{NR}} & {\small \color{MyDarkGray} \emph{NR}} & {\color{MyDarkGray}57} & {\small \color{MyDarkGray} \emph{NR}} & {\small \color{MyDarkGray} \emph{NR}} & 62 & \textbf{27} \\
  (Label Noise) & 85.0\% & {\small \color{MyDarkGray} \emph{NR}} {\color{MyDarkGray} (28\%)} & {\small \color{MyDarkGray} \emph{NR}} {\color{MyDarkGray} (23\%)} & {\small \color{MyDarkGray} \emph{NR}} {\color{MyDarkGray} (84\%)} & {\small \color{MyDarkGray} \emph{NR}} {\color{MyDarkGray} (48\%)} & {\small \color{MyDarkGray} \emph{NR}} {\color{MyDarkGray} (62\%)} & {\small \emph{NR}} (85\%) & \textbf{49 (91\%)} \\

  \rowcolor{MyLightGray} CIFAR100$^*$ & 40.0\% & {\color{MyDarkGray}138} & {\color{MyDarkGray}139} & {\color{MyDarkGray}71} & {\small \color{MyDarkGray} \emph{NR}} & {\color{MyDarkGray}93} & 65 & \textbf{48} \\
  \rowcolor{MyLightGray} & 52.5\% & {\small \color{MyDarkGray} \emph{NR}} {\color{MyDarkGray} (42\%)} & {\small \color{MyDarkGray} \emph{NR}} {\color{MyDarkGray} (42\%)} & {\color{MyDarkGray}132} {\color{MyDarkGray} (55\%)} & {\small \color{MyDarkGray} \emph{NR}} {\color{MyDarkGray} (18\%)} & {\small \color{MyDarkGray} \emph{NR}} {\color{MyDarkGray} (43\%)} & 133 (54\%) & \textbf{77 (61\%)} \\

  CIFAR100$^*$ & 40.0\% & {\small \color{MyDarkGray} \emph{NR}} & {\small \color{MyDarkGray} \emph{NR}} & {\color{MyDarkGray}94} & {\small \color{MyDarkGray} \emph{NR}} & {\color{MyDarkGray}89} & 79 & \textbf{49} \\
  (Label Noise) & 47.5\% & {\small \color{MyDarkGray} \emph{NR}} {\color{MyDarkGray} (4\%)} & {\small \color{MyDarkGray} \emph{NR}} {\color{MyDarkGray} (4\%)} & {\color{MyDarkGray}142} {\color{MyDarkGray} (48\%)} & {\small \color{MyDarkGray} \emph{NR}} {\color{MyDarkGray} (14\%)} & {\small \color{MyDarkGray} \emph{NR}} {\color{MyDarkGray} (43\%)} & 116 (50\%) & \textbf{65 (60\%)} \\

  \rowcolor{MyLightGray} CINIC10 & 70.0\% & {\small \color{MyDarkGray} \emph{NR}} & {\small \color{MyDarkGray} \emph{NR}} & {\color{MyDarkGray}34} & {\small \color{MyDarkGray} \emph{NR}} & {\small \color{MyDarkGray} \emph{NR}} & 38 & \textbf{27} \\
  \rowcolor{MyLightGray} & 77.5\% & {\small \color{MyDarkGray} \emph{NR}} {\color{MyDarkGray} (36\%)} & {\small \color{MyDarkGray} \emph{NR}} {\color{MyDarkGray} (50\%)} & {\color{MyDarkGray}64} {\color{MyDarkGray} (82\%)} & {\small \color{MyDarkGray} \emph{NR}} {\color{MyDarkGray} (39\%)} & {\small \color{MyDarkGray} \emph{NR}} {\color{MyDarkGray} (60\%)} & 97 (80\%) & \textbf{38 (83\%)} \\

  CINIC10 & 60.0\% & {\small \color{MyDarkGray} \emph{NR}} & {\small \color{MyDarkGray} \emph{NR}} & {\color{MyDarkGray}22} & {\small \color{MyDarkGray} \emph{NR}} & {\color{MyDarkGray}30} & 24 & \textbf{13 }\\
  (Label Noise) & 67.5\% & {\small \color{MyDarkGray} \emph{NR}} {\color{MyDarkGray} (16\%)} & {\small \color{MyDarkGray} \emph{NR}} {\color{MyDarkGray} (16\%)} & {\color{MyDarkGray}35} {\color{MyDarkGray} (79\%)} & {\small \color{MyDarkGray} \emph{NR}} {\color{MyDarkGray} (39\%)} & {\small \color{MyDarkGray} \emph{NR}} {\color{MyDarkGray} (64\%)} & 38 (78\%) & \textbf{17 (82\%)} \\

  \rowcolor{MyLightGray} SST2$^*$ & 82.5\% & {\small \color{MyDarkGray}8} & {\small \color{MyDarkGray}2} & {\color{MyDarkGray}3} & {\small \color{MyDarkGray} \emph{NR}} & {\small \color{MyDarkGray}7} & \textbf{1} & \textbf{1} \\
  \rowcolor{MyLightGray} & 90.0\% & {\small \color{MyDarkGray} \emph{NR}} {\color{MyDarkGray} (87\%)} & {\small \color{MyDarkGray}4} {\color{MyDarkGray} (91\%)} & {\color{MyDarkGray} \emph{NR}} {\color{MyDarkGray} (89.7\%)} & {\small \color{MyDarkGray} \emph{NR}} {\color{MyDarkGray} (66\%)} & {\small \color{MyDarkGray} \emph{NR}} {\color{MyDarkGray} (83\%)} & 6 (90\%) & \textbf{3 (92\%)} \\

  CoLA$^*$ & 75.0\% & {\small \color{MyDarkGray}8} & {\small \color{MyDarkGray}6} & {\color{MyDarkGray}16} & {\small \color{MyDarkGray} \emph{NR}} & {\color{MyDarkGray} \emph{NR}} & 34 & \textbf{3}\\
  & 80.0\% & {\small \color{MyDarkGray} \emph{NR}} {\color{MyDarkGray} (78\%)} & {\small \color{MyDarkGray} \emph{NR}} {\color{MyDarkGray} (79\%)} & {\color{MyDarkGray} \emph{NR}} {\color{MyDarkGray} (78\%)} & {\small \color{MyDarkGray} \emph{NR}} {\color{MyDarkGray} (62\%)} & {\small \color{MyDarkGray} \emph{NR}} {\color{MyDarkGray} (69\%)} & \emph{NR} (76\%) & \textbf{39 (80\%)} \\

 \hline \end{tabular}
  \label{tab:speedups}
\end{adjustbox}
\end{table*}

\subsection{Properties of RHO-LOSS \& Other Selection Functions}
\label{sec:properties}

We established that RHO-LOSS can accelerate the training of various target architectures with a single IL model, even if the IL model is smaller and has considerably lower accuracy than the target models (Section~\ref{sec:cheap_irlo}). This suggests robustness to target-IL architecture mismatches.

To understand this robustness, we investigate the properties of points selected by RHO-LOSS, when the target and IL model architectures are identical, and when the IL model is smaller. Explaining the robustness, we find that, in both cases, RHO-LOSS prioritizes points that are \textcolor{myblue}{non-noisy}, \textcolor{mygreen}{task-relevant}, and \textcolor{myred}{not redundant}. 
We also investigate the properties of points selected by prior art. 

\paragraph{Noisy points.} We investigate how often different methods select noisy points by uniformly corrupting the labels for 10\% of points and tracking what proportion of selected points are corrupted. RHO-LOSS deprioritizes noisy points for both IL models (Fig.~\ref{fig:properties}). We observe a failure mode of the widely-used loss and gradient norm selection functions: they 
select far more noisy points than uniform. These methods also severely drop in accuracy when the noise follows the class confusion matrix \citep{rolnick2017deep} and when we add ambiguous images \citep{mukhoti2021deterministic} (Appendix~\ref{app:noise}).

Together, this suggests that noisy points have high loss (and gradient norm), but also high IL and thus low reducible loss. Their IL is high even when the IL model is small as noisy labels cannot be predicted well using the holdout set.

\paragraph{Relevant points.} We study how often less relevant points are selected by creating the CIFAR100 Relevance dataset, in which 80\% of the data comes from 20\% of the classes. 
This mimics natural distributions of NLP and vision data where most data comes from few object classes, topics, or words \citep{baayen2001word,tian2021divide}. 
Concretely, we retain all examples from 20 randomly chosen ``high relevance'' classes but only 6\% of the examples from other, ``low relevance'' classes. Intuitively, since the high relevance classes have higher $\ptof{x}$ and are 17x more likely to appear at test time, improving their accuracy improves the test accuracy much more than improving the accuracy of less relevant classes. 

The loss and gradient norm methods select more points than uniform selection from the low relevance classes (Fig.~\ref{fig:properties}).
In contrast, RHO-LOSS selects somewhat fewer low relevance points, suggesting these classes have high IL.
Since the less relevant classes are less abundant in the holdout set, both the small and large IL models have higher loss on them.

\paragraph{Redundant points.} To investigate whether methods select redundant points, we track the percentage of selected points that are already classified correctly. This is only a proxy for redundancy; points that are classified correctly but with low confidence are not fully redundant, since their loss can be further reduced. We control for the different accuracy reached by each method by averaging only over epochs in which test accuracy is lower than the final accuracy reached by the weakest performing method. Fig.~\ref{fig:properties} shows that all methods select fewer redundant points than uniform sampling.

\subsection{Speedup}
\label{sec:speedups}
Finally, we evaluate how much different selection methods speed up training. 
Recall that the main application area for our work is large web-scraped datasets, known for high levels of noise and redundancy. Clothing-1M is such a dataset (Section~\ref{sec:experiments}.0). We also include smaller, clean benchmarks from vision (CIFAR-10, CIFAR-100, CINIC-10) and NLP (CoLA, SST-2). 
Finally, we study if selection functions are robust to the controlled addition of label noise.

\paragraph{Speedup on clean data.}
RHO-LOSS reaches target accuracies in fewer epochs than uniform selection on all datasets (Table~\ref{tab:speedups}). It also outperforms state-of-the-art methods by a clear margin in terms of speed and final accuracy. On the challenging CoLA language understanding dataset, the speedup over uniform selection exceeds $10$x.
In Table~\ref{tab:double_irlomo} (Appendix~\ref{app:steps_needed}), we find similar speedups when using no holdout data. 

\paragraph{Speedup on noisy data.}
When adding $10$\% label noise, batch selection with RHO-LOSS achieves greater speedups while, as hypothesized, prior art degrades (Table~\ref{tab:speedups}). Notably, on noisier data, the speedup over uniform selection grows. 

\paragraph{Speedup on large web-scraped data.} 
On Clothing-1M, loss-based and gradient norm-based selection fail to match uniform selection, suggesting they are not robust to noise. In contrast, RHO-LOSS reaches the highest accuracy that uniform selection achieves during 50 epochs in just 2 epochs and improves final accuracy (72\% vs 70\%). Notably, this was possible even though the IL model we used has low accuracy (62.2\%) and was trained on ca. $10$x less data. RHO-LOSS also used 2.7x fewer FLOPs to reach the peak accuracy of uniform selection, including the cost of training the IL model (which could be amortized) and despite our implementation being designed to save time, not compute. While Table~\ref{tab:speedups} shows results for a Resnet-50, Fig.~\ref{fig:clothing-1M} includes additional architectures, with an average speedup of $18$x.

\section{Related Work}
\label{sec:related}

\paragraph{Time-efficient data selection.}
Forward passes for selection can be accelerated using low-precision hardware or parallelization. While backward passes typically require high precision, forward passes can tolerate lower precision \citep{jouppi2017datacenter, jiang2019accelerating}, especially as we only need the loss (not the activations which would be needed for backprop). A forward pass by default requires roughly $3$x less time than a forward-backward pass but this speedup can be increased to a factor around $10$x when using the low-precision cores available in modern GPUs and TPUs \citep{jouppi2017datacenter, jiang2019accelerating}. Further, prior work uses a set of workers that perform forward passes on $B_t$ or on the entire dataset asynchronously while the master process trains on recently selected data \citep{alain2015variance}. 

\paragraph{Compute-efficient data selection.}
While we limit our scope to comparing selection functions and we compute them naively, this choice is inefficient in practice. Selection can be made cheaper by reusing losses computed in previous epochs \citep{loshchilov2015online, jiang2019accelerating} or training a small model to predict them \citep{katharopoulos2017biased, zhang2019autoassist, coleman2019selection}. Alternatively, core set methods perform selection once before training \citep{mirzasoleiman2020coresets, borsos2020coresets}, although typically with more expensive selection functions.

\paragraph{Data selection functions.}
RHO-LOSS is best understood as an alternative to existing selection functions, which can be categorized by the properties of points they select and whether they use information about labels. ``Hard’’ points are selected both by high loss \citep{loshchilov2015online,kawaguchi2020ordered,jiang2019accelerating} and high prediction uncertainty \citep{settles2009active, li2006confidence, coleman2019selection}. However, prediction uncertainty does not require labels and can thus be used for active learning.
Despite this, they both suffer from the same problem: high loss and high uncertainty can be caused by noisy (in particular, ambiguous) labels.
This also applies to selection of points whose labels are easily forgotten during training \citep{toneva2018empirical}. Noisy points are avoided by our negative IL baseline and comparable offline selection methods \citep{pleiss2020identifying, chen2019understanding, paul2021deep}.
Points that most reduce (expected) holdout loss are also selected for other purposes \citep{kirsch2021evalbald, killamsetty2020glister, ren2018learning}, although using much more computation.

\paragraph{Variance reduction methods.}
Online batch selection is also used to reduce the variance of the gradient estimator computed by SGD \citep{katharopoulos2018not, katharopoulos2017biased,johnson2018training,alain2015variance}, which is widely used in reinforcement learning \citep{schaul2015prioritized}. Such methods typically use importance sampling---points with high (approximate) gradient norm are sampled with high probability but then down-weighted in the gradient calculation to de-bias the gradient estimate. 
Without de-biasing, methods like RHO-LOSS also create selection bias. However, bias can improve test performance, both in theory and practice \citep{farquhar2021statistical, kawaguchi2020ordered}.


\section{Conclusion}

To reduce excessive training times, we introduce a theoretically grounded selection function that enables substantial speedups on clean data and even larger speedups on noisy and web-scraped data. 
By illuminating three properties of optimal selection, we hope to motivate new directions in batch selection. 
However, our selection function should be combined with methods in Section~\ref{sec:related} for cheap and fast selection with maximal speedups.

\section*{Ethics Statement}

It will be important to understand how subset selection might affect performance on data about minority groups. The selection may prioritize rare groups since majority groups are learnt more quickly, or deprioritizes rare groups since they affect the loss on holdout data less. 

Since such biases can also stem from the dataset itself \citep{mehrabi2021survey}, it should be investigated if our method can remove data biases through the use of an unbiased holdout set. By training the irreducible loss model on unbiased data, we can implicitly specify that the model should perform well on unbiased data, even when the training set contains bias. This may be useful for specifying that all groups are equally important to learn.

\section*{Acknowledgements} For useful feedback we thank Pascal Notin and Kelsey Doerksen.

\section*{Author Contributions}
S\"{o}ren Mindermann, Jan Brauner, Mrinank Sharma and Muhammed Razzak designed and analysed the experiments shown in the paper. 

Jan Brauner implemented experiments on ALBERT, CIFAR-10, experiments in Figure 2 and 7, Table 3 and 4, among others.
Muhammed Razzak implemented experiments on Clothing-1M, CIFAR-100, and MNIST, among others. 
Mrinank Sharma implemented experiments on CINIC-10, experiments in Figures 3 and 9, among others. 

S\"{o}ren Mindermann and Muhammed Razzak implemented pilot experiments. 

Winnie Xu and Adrien Morisot implemented early experiments on language models, advised by Aidan Gomez. 

S\"{o}ren Mindermann, Jan Brauner, Mrinank Sharma, Muhammed Razzak, Benedikt H\"{o}ltgen and Andreas Kirsch wrote the paper.

S\"{o}ren Mindermann conceived of the algorithm.

S\"{o}ren Mindermann, Jan Brauner, Andreas Kirsch and Yarin Gal developed the theory. 

S\"{o}ren Mindermann led and managed the research.

Yarin Gal and Sebastian Farquhar advised the research.

\bibliographystyle{icml2022}
\bibliography{references}

\newpage
\clearpage
\onecolumn
\section*{\Large Appendix}
\bigskip
\appendix

\section{Steps required to reach a given test accuracy} \label{app:steps_needed}

\begin{figure*}[h]
    \centering
    \includegraphics[width=\textwidth]{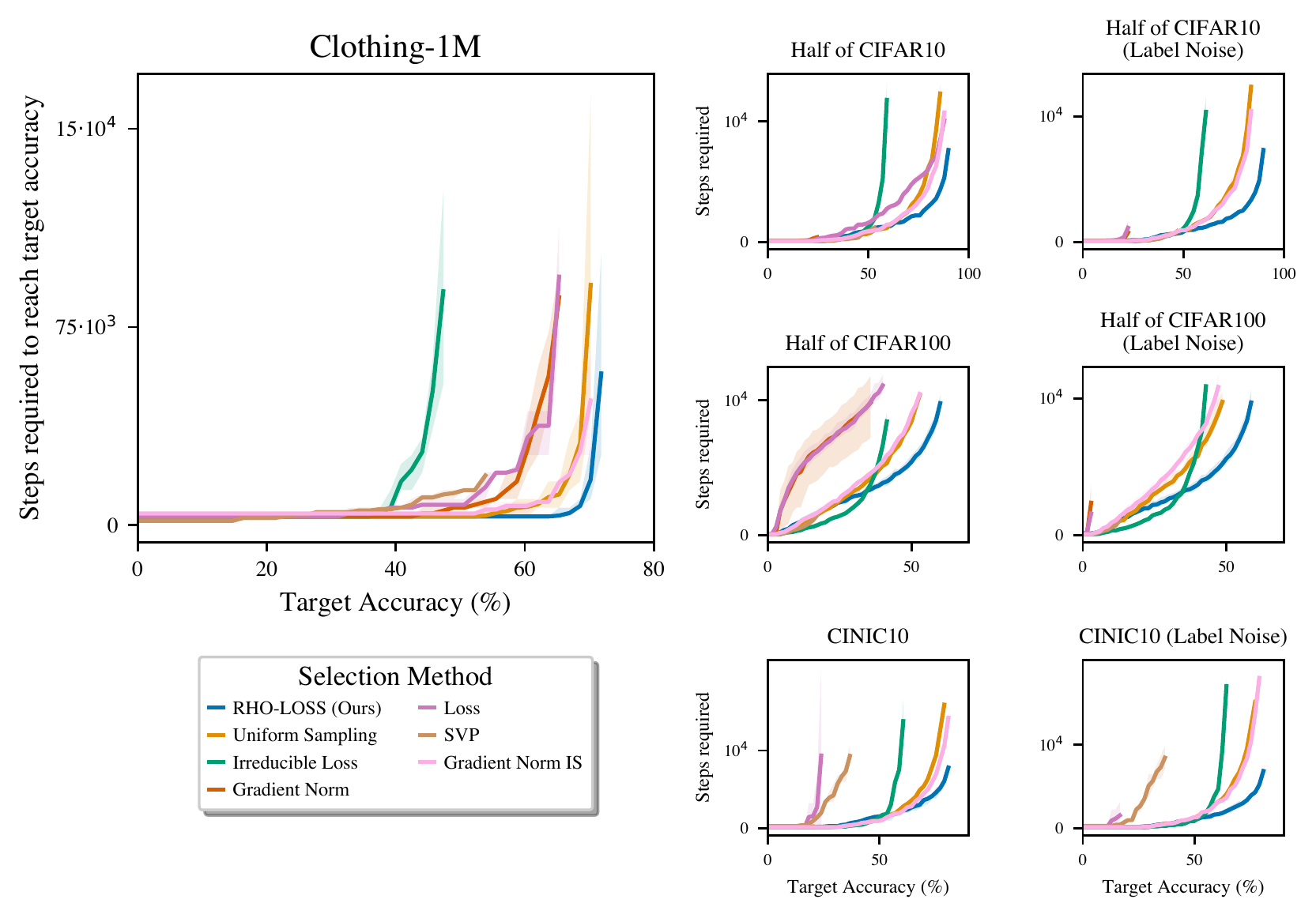}
    \vspace{-2mm}
    \caption{Vision datasets---gradient steps required to achieve a given test accuracy \textbf{(lower is better)}. \textbf{Left column:} The speedup of RHO-LOSS over uniform sampling is greatest on a large-scale web-scraped dataset with noisy labels.  \textbf{Middle column:}  Speedups are still substantial on clean datasets and RHO-LOSS still achieves higher final accuracy than all prior art. \textbf{Right column:} Applying $10\%$ uniform label noise to training data degrades other methods but increases the speedup of our method. A step corresponds to lines $5-10$ in Algorithm~1. Lines correspond to means and shaded areas to minima and maxima across 3 random seeds. On CIFAR10/100, only half of the data is used for training (see text).}
    \label{fig:walltime}
\end{figure*}

\begin{figure*}
    \centering
    \includegraphics[width=\textwidth]{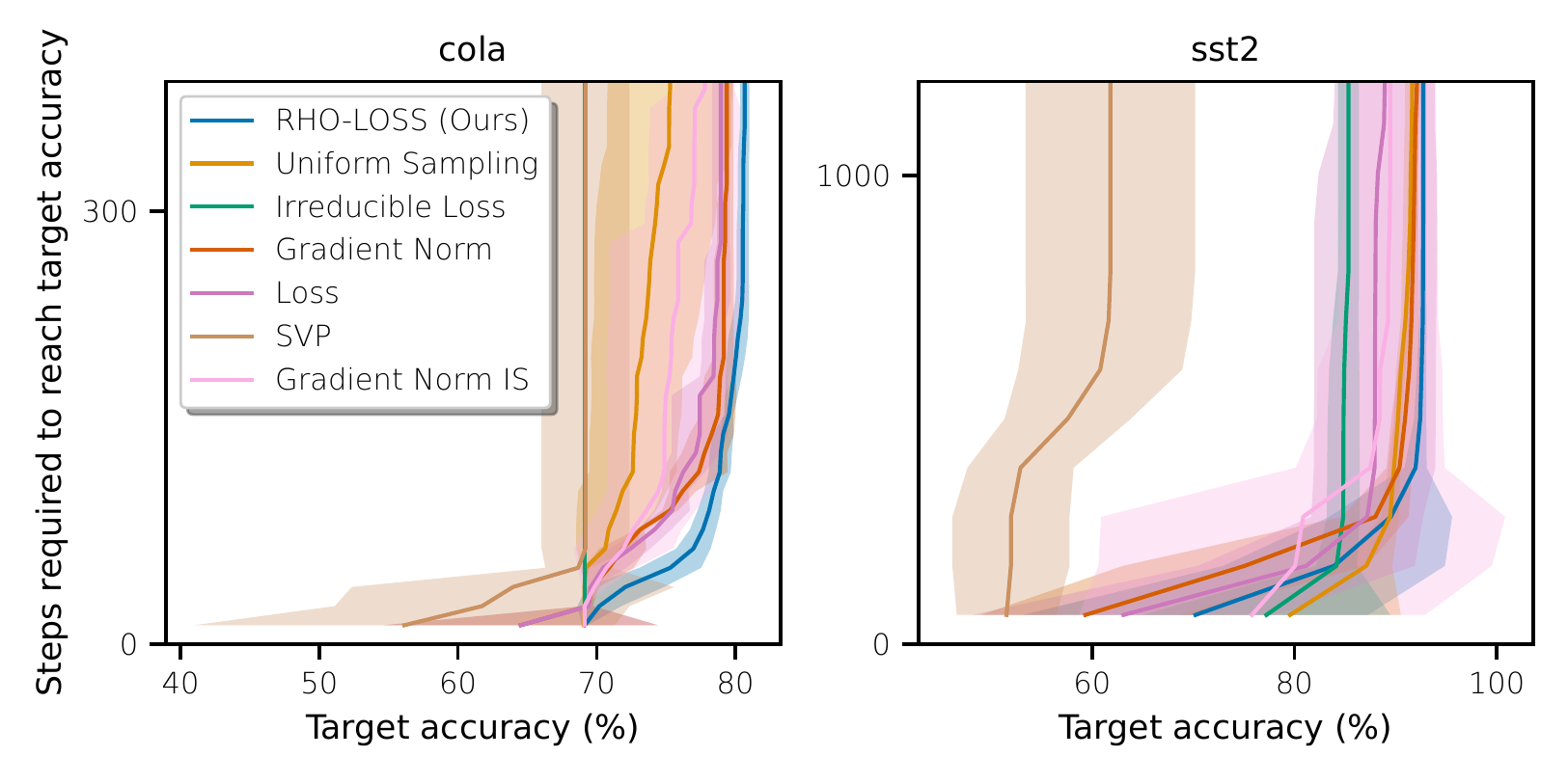}
    \vspace{-2mm}
    \caption{NLP datasets---gradient steps required to achieve a given test accuracy \textbf{(lower is better)}. \textbf{Left:} CoLA grammatical acceptibility classification. \textbf{Right:} SST2 sentiment classification. A step corresponds to lines $5-10$ in Algorithm~1. Lines correspond to means and shaded areas to standard deviations across 4 or more random seeds. Only half of the data is used for training (see text).}
    \label{fig:walltime_nlp}
\end{figure*}

Figs.~\ref{fig:walltime}~(vision)~and~\ref{fig:walltime_nlp}~(NLP) show the number of steps required to reach a given test accuracy across several datasets for different selection methods. Interestingly, on CoLA (unbalanced and noisy), the uniform sampling baseline shows high variance across seeds, while RHO-LOSS works robustly across seeds. 

Table~\ref{tab:double_irlomo} shows results for RHO-LOSS training without holdout data. Results are similar to Table~\ref{tab:speedups}. Here, we train the IL model without any holdout data. We split the training set $\D$ into two halves and train an IL model on each half.  Each model computes the IL for the half of $\D$ that it was not trained on. (This is as in Fig.~\ref{fig5}, row 3, except that previously we only used half of $\D$ and further split it into halves of the half.) Training two IL models costs no additional compute since each model is trained on half as much data compared to the default settings.

\begin{table*}[]
\caption{\textbf{Epochs required to reach a given target test accuracy when using no holdout data} (lower is better). Final accuracy in parentheses. Results averaged across 2-3 seeds. Best performance in bold. RHO-LOSS performs best in both epochs required and final accuracy.}
\begin{tabular}{@{}c|c|cc@{}}
Dataset               & Target Acc & Uniform      & RHO-LOSS              \\ \midrule
CIFAR10               & 80\%       & 39           & \textbf{17}           \\
\multicolumn{1}{l|}{} & 90\%       & 177 (90.8\%) & \textbf{47 (92.2\%)}  \\ \midrule
CIFAR100              & 50\%       & 47           & \textbf{22}           \\
\multicolumn{1}{l|}{} & 65\%       & 142 (67.8\%) & \textbf{87 (68.1\%)}  \\ \midrule
CINIC10               & 70\%       & 37           & \textbf{26}           \\
\multicolumn{1}{l|}{} & 80\%       & 146 (80.1\%)  & \textbf{70 (82.1\%)}
\end{tabular}
\label{tab:double_irlomo}
\end{table*}

\FloatBarrier
\section{Experiment Details}
\label{app:exp_details}
\paragraph{Architectures.} We experiment with various architectures in Figs.~\ref{fig:clothing-1M} and \ref{fig5} (row 4). In all other figures and tables, we use the following architectures: For experiments on QMNIST, we use a multi-layer perceptron with 2 hidden layers and 512 units in each hidden layer. For experiments on CIFAR-10, CIFAR-100 and CINIC-10, we use a variant of ResNet-18 \citep{he2016deep}. We adapted the ResNet18 to 32x32 images by modifying the architecture to remove the downsampling effect. We replaced the spatial downsampling of a strided convolution and max pooling in the original ResNet18, with a convolutional layer with 64 filters and a kernel size of 3x3. We also removed the average pooling at the end of the ResNet18. This ResNet18 variant is similar to Resnet20, just with more filters. For experiments on Clothing-1M, following the experimental set-up of \citet{Yi_2019_CVPR}, the target model is a ResNet-50 pre-trained on ImageNet. The irreducible loss model is a ResNet-18 with random initialisation. The multiple target architectures in Fig~\ref{fig5} were adapted from \cite{huyphan_2021}. For NLP datasets, we use a pretrained ALBERT v2 \citep{lan2019albert}.

\paragraph{Hyperparameters.} \textit{Vision}: All models are trained using the AdamW optimizer with default PyTorch hyperparameters ($\beta_1$=0.9, $\beta_2$=0.999, and weight decay of $0.01$, learning rate $0.001$), a $n_b=32$ ($64$ for CINIC-10) $n_B=320$ ($640$ for CINIC-10), meaning we select $\frac{n_b}{n_B}=10\%$ of points. \textit{NLP}: ALBERT v2 was trained using the AdamW optimizer with a learning rate as indicated in the original paper ($2 \cdot 10^{-5})$ and weight decay of 0.02. We finetuned all weights, not just the final layer. the batch size $n_b$ was $32$, $n_B=320$, meaning we select $\frac{n_b}{n_B}=10\%$ of points. We use between 2 and 10 seeds for each experiment.

\paragraph{Data augmentation.} On CIFAR-10, CIFAR-100, and CINIC-10, we train using data augmentation (random crop and horizontal flip), both for training the IL model, and in the main training runs. Remember that we only compute the irreducible losses once at the start of training, to save compute (Algorithm~\ref{alg:RHOLOSS}). We use the un-augmented images for this as we found that using augmented images makes little difference to performance but costs more compute.

\paragraph{Irreducible loss model training.} The irreducible loss models are trained on holdout sets (not test sets, see dataset description in main text). For each dataset, we select the irreducible loss model checkpoint from the epoch with \textit{lowest holdout loss} on $\D$ (as opposed to highest accuracy); we find that this improves performance while also saving compute as the holdout loss typically reaches its minimum early in training.

\paragraph{BatchNorm.} Like many deep-learning methods, RHO-LOSS interacts with BatchNorm \cite{ioffe2015batch} since the loss of a given point is affected by other points in the same batch. \textbf{Important:} We compute the BatchNorm statistics for selection and model update separately. For selection (line 5-8 in Algorithm~\ref{alg:RHOLOSS}), the statistics are computed across the large batch $B_t$. For training (line 9-10), the statistics are computed across the small batch $b_t$. These choices can affect performance a lot. For new datasets, we recommend to vary how the batchnorm statistics are computed during selection (trying both train mode and eval mode) and choose the option that works best.

\FloatBarrier

\section{Robustness to Noise} \label{app:noise}
\begin{figure*}[ht]
    \centering
    \includegraphics[width=\textwidth]{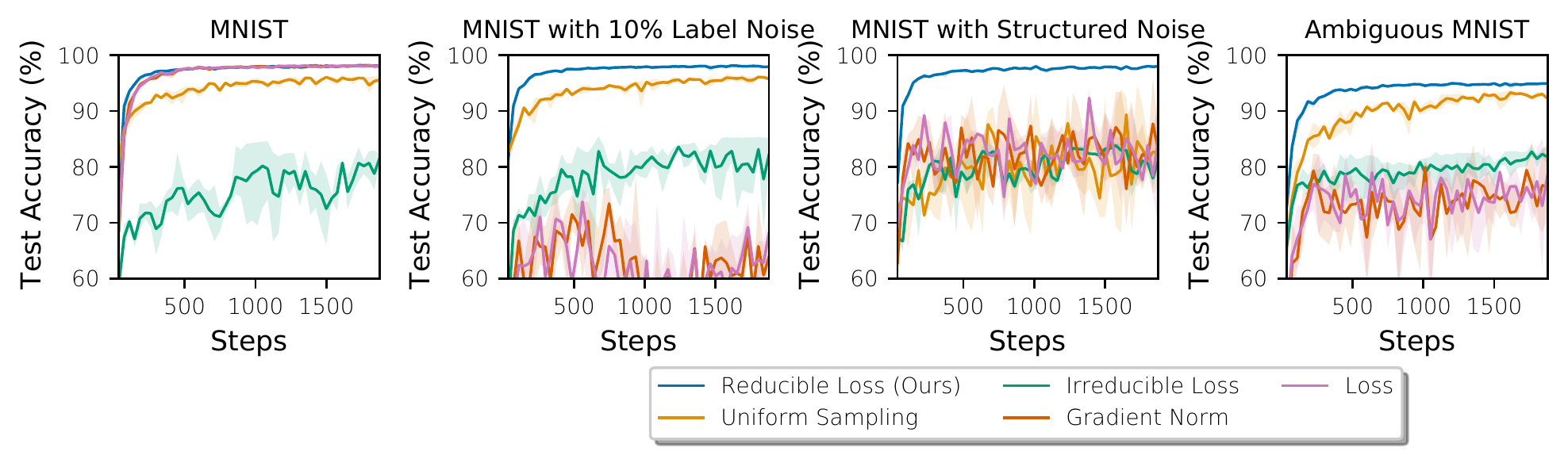}
    \caption{RHO-LOSS is robust to a variety of label noise patterns, while other selection methods degrade. A step corresponds to lines $6-11$ in Algorithm~\ref{alg:RHOLOSS}. Lines correspond to means and shaded areas to minima and maxima across 3 random seeds.}
    \label{fig:mnist}
\end{figure*}

In this set of experiments, we evaluate the performance of different selection methods under a variety of noise patterns on QMNIST (MNIST with extra holdout data) and variations thereof. We use this dataset because it has little label noise in its original form, allowing us to test the effect of adding noise. Firstly, we add uniform label noise to 10\% of training points. Secondly, we add structured label noise that affects easily confused classes. We follow \cite{rolnick2017deep} and flip the labels of the four most frequently confused classes (in the confusion matrix of a trained model) with 50\% probability. For example, a 2 is often confused with a 5; thus we change the label of all 2s to 5s with 50\% probability. Thirdly, we leverage the natural noise distribution of MNIST by using AmbiguousMNIST \citep{mukhoti2021deterministic} as the training set. AmbiguousMNIST contains a training set with 60k generated ambiguous digits that have more than one plausible label. While selecting with loss and gradient norm trains accelerates training on the MNIST training set, their performance degrades on all three types of noise distributions (Figure \ref{fig:mnist}).

\FloatBarrier
\section{Irreducible Holdout Loss Approximation}\label{app:update_irreducible_2}

In this appendix section, we examine one of the key approximations made in the theory section. To arrive at Eq.~(\ref{eq:rholoss}), we used the approximation $\Lof{y \given x; \Dval} \approx \Lof{y \given x; \Dval, \Dtrain}$. In words, we approximated the cross-entropy loss of a model trained on the data points acquired so far $\Dtrain$ and the holdout dataset $\Dval$, with the cross-entropy loss of a model trained only on the holdout set. This approximation saves a lot of compute: rather than having to recompute the term with every change of $\mathcal{D}_t$, it is now sufficient to compute it once at the start of training.

We have already highlighted the impact of the approximation on points selected when training on QMNIST in Section~\ref{sec:impact_of_approximations}. In our main experiment setting---using neural networks trained with gradient descent---we empirically find that the approximation does not reduce speed of target model training or final target model accuracy (Table~\ref{tab:update_irreducible_approximation}). This finding holds across a range of datasets (CIFAR-10, CIFAR-100, CINIC-10). Updating the irreducible loss model on $\Dtrain$ seems empirically not necessary.

\begin{table}[]
\setlength{\tabcolsep}{2pt}
\caption{Number of epochs required to reach a given target test accuracy across several datasets. Results averaged across 2-3 random seeds. NR indicates that the target accuracy was not reached.}
\centering
\begin{tabular}{@{}cccc@{}}
\toprule
Dataset  & Target acc                & approximated selection function & original selection function        \\
& & $\Lof{y \given x; \Dtrain} - \Lof{y \given x; \Dval}$ & $\Lof{y \given x; \Dtrain} - \Lof{y \given x; \Dval, \Dtrain}$ \\ \midrule
  & 60\% & 18                              & 13                                 \\
       CIFAR10 & 75\% & 30                              & 24                                 \\
        & 90\% & 102                             & NR, but reaches 88\% in 157 epochs \\ \midrule
 & 30\% & 35                              & 21                                 \\
    CIFAR100    & 45\% & 58                              & NR, but reaches 43\% in 61 epochs  \\
        & 60\% & 123                             & NR                                 \\ \midrule
       & 55\% & 12                              & 12                                 \\
     CINIC10    & 65\% & 19                              & 21                                 \\
        & 75\% & 32                              & NR, but reaches 74\% in 68 epochs  \\ \bottomrule
\end{tabular}

\label{tab:update_irreducible_approximation}
\end{table}

Indeed, the approximation actually has two desirable properties when used for neural networks trained with gradient descent. We will first describe why we expect these desirable properties, and then show that they indeed appear. 

First, let us restate both selection functions:

Original selection function:\phantom{Approximated selection function} 
\begin{align*}
\hspace{0.3\linewidth}&\hspace{0.7\linewidth}\\[-1.35\baselineskip]
    \argmax_{(x,y) \in B_t}\ &\Lof{y \given x; \Dtrain} - \Lof{y \given x; \Dval, \Dtrain}.\\[-1.35\baselineskip]
\end{align*}
Approximated selection function:\phantom{Original selection function} 
\begin{align*}
\hspace{0.3\linewidth}&\hspace{0.7\linewidth}\\[-1.35\baselineskip]
\argmax_{(x,y) \in B_t}\ &\Lof{y \given x; \Dtrain} - \Lof{y \given x; \Dval}.\\[-1.35\baselineskip]
\end{align*}

\paragraph{Desirable property 1: The approximation prevents repeated selection of undesirable points.} When using SGD instead of Bayesian updating, the original selection function can acquire undesired points repeatedly. Let's say that we acquire, for whatever reason, a noisy, redundant, or irrelevant point. We only take one gradient step each time we acquire a (batch of) point(s), meaning the training loss (first term in the selection function) will on each only decrease somewhat. In the original selection function, the second term will also decrease somewhat, meaning that the difference between the first and second term may remain large. In the approximated selection function, the second term is constant, the difference between first and second term will thus likely decrease more than under the original selection function. Under the approximated selection function, we are thus less likely to acquire undesired points again, if we have acquired them in earlier epochs.

\paragraph{Desirable property 2: The approximation prevents deterioration of the irreducible loss model over time.} With both selection functions, we compute the second term of the selection function with an "irreducible loss model", which we train on a holdout set before we start target model training. In the target model training, we (greedily) acquire the points that most improve the loss of the target model (on the holdout set). We thus deliberately introduce bias into the data selection. However, this bias is tailored to the target model and may not be suitable for the irreducible loss model. As a simplifying example, consider a target model early in training, which has not yet learnt a certain class, and an irreducible loss model, which has learnt that class. Data points in that class will have high training loss, low irreducible loss, and will be acquired often. This, however, is not useful for the irreducible loss model, and might lead to decreased accuracy on data points from other classes. With the approximation, this can't happen. The described failure mode could likely also be alleviated by more sophisticated training schemes for the irreducible loss model, such as periodically mixing in data points from the holdout set. However, such training schemes would require even more compute and/or overhead.

\vspace{-2mm}
\begin{figure}
    \centering
    \hspace*{-9mm}\includegraphics[width=1\linewidth]{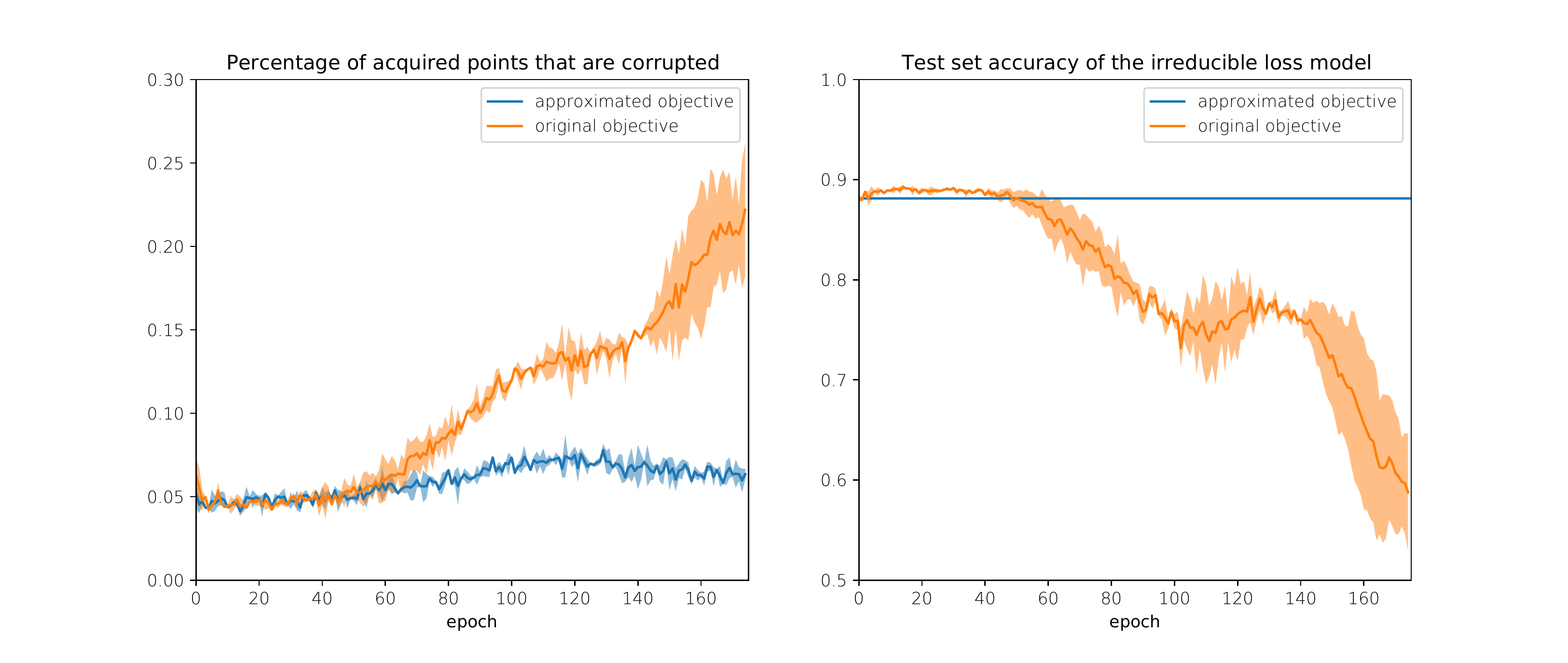}
    \caption{Desired properties of the irreducible loss model approximation. \textbf{Left.} The approximated selection function selects fewer corrupted points later on in training. \textbf{Right.} The test set accuracy of the irreducible loss model deteriorates over time if it is updated on $\mathcal{D}_t$. With the approximation, the irreducible loss is not updated during target model training. Results on CIFAR-10 with 20\% of data points corrupted with uniform label noise. Shaded areas represent standard deviation across three different random seeds. }
    \label{fig:update_irlomo_failure_modes}
\end{figure}

We find empirically that both desired properties of the approximation indeed manifest themselves. In Fig.~\ref{fig:update_irlomo_failure_modes}, we train a target model (Resnet-18) on CIFAR-10, with 20\% of the data points corrupted by uniform label noise. The approximated selection function leads to faster target model training (the approximated selection function needs 80 epochs to reach the same target model accuracy that the original selection function reaches in 100 epochs) and higher final accuracy than the original selection function (88.6\% vs 86.1\%). Indeed, the original selection function leads to acquiring more corrupted points, especially later in training (Fig.~\ref{fig:update_irlomo_failure_modes}, left), and the accuracy \textit{of the irreducible loss model} deteriorates over time (Fig.~\ref{fig:update_irlomo_failure_modes}, right). We tuned the learning rate of the irreducible loss model to 0.01 times that of the target model. Without this adjustment, the results look similar but the original selection function performs worse.

\FloatBarrier
\section{Experimental Details for Assessing Impact of Approximations}\label{sec:approximation_appendix}
\textbf{Dataset:} QMNIST, with uniform label noise applied to 10\% of the dataset. Batch size of 1000 is used.

\textbf{Models:} Deep Ensemble contains 5 3-layer MLP's with 512 hidden units. The weaker irreducible loss model is an MLP with 256 hidden units.

\textbf{Training:} 
For Approximation 0, we use a deep ensemble for both models. The irreducible loss model is trained to convergence on $\Dval$. Then the target model and the irreducible model are used to acquire 10\% of points each batch using the selection function. They are then trained to convergence on each batch of points acquired. The irreducible loss model is trained on $\Dval \cup \Dtrain$, while the target model is only trained on $\Dtrain$. We train for a maximum of 5 epochs, which often is to convergence, to enable a fair comparison to further approximations.
For Approximation 1a, the deep ensembles are replaced with single MLPs. The training regime remains the same. We compare the approximations over the first epoch.
To compare Approximation 1b to 0, and for all further approximations, we increase the size of the dataset five-fold, by duplicating samples in QMNIST. This means for approximation 1b, we have 5x the data that we have for Approximation 1a, but with increased redundancy. We train the model in Approximation 1b by taking a single gradient step per datapoint, with the larger dataset. On the other hand, we train the model for Approximation 0 (still to convergence or 5 epochs) on the standard dataset size. By doing this, Approximation 0 and 1b have taken the equivalent number of gradient steps, at the time-steps where we are tracking the reducible loss of points selected, enabling a fair comparison between the approximations. The irreducible loss models are trained on $\Dval \cup \Dtrain$ in their respective set-ups.  
To compare Approximation 2 to Approximation 0, we compare updating the irreducible loss model with a single gradient on each set of acquired points, to not updating the irreducible loss model on $\Dtrain$ at all. To isolate effect of not updating, we utilise the same initial irreducible loss model.
To compare Approximation 3, we simply train a small irreducible model (one with 256 hidden units) and follow the same training regime as Approximation 2.

\FloatBarrier

\FloatBarrier
\section{Ablation of percentage selected}
\label{app:selectivity}
\begin{figure}[h]
    \centering
    \includegraphics[width=0.5\linewidth]{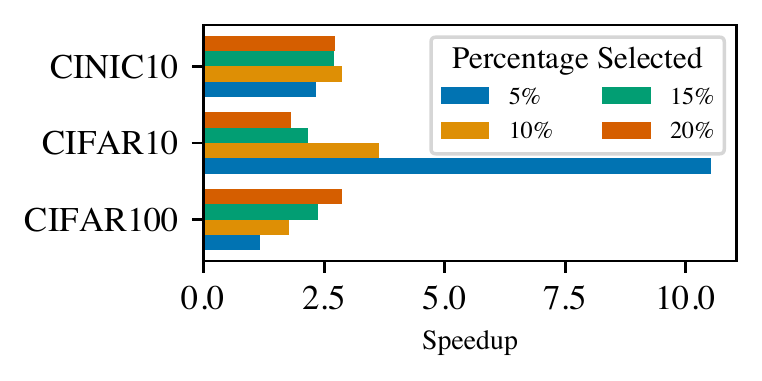}
    \caption{Varying the percent of data points selected in each training batch. Average over 3 random seeds.}
    \label{fig:percent_train}
\end{figure}

Our method has a hyperparameter, the percentage $\frac{n_b}{n_B}$ of evaluated points which are selected for training. In the experiments above, this parameter was set to $0.1$. We have not tuned this parameter, as we aim to analyse how well our method works ``out of the box". In fact, on 2/3 datasets, performance further improves with other values of this parameter. Adjusting this percentage should allow practitioners to specify their preferred tradeoff between training time and computation, where a low percentage typically corresponds to a lower training time and greater compute cost. For these experiments, we kept $n_b=32$ and adapt $n_B$ accordingly. The percentage  $\frac{n_b}{n_B}$ of datapoints selected per batch has different effects across datasets as shown in Fig.~\ref{fig:percent_train}.

\FloatBarrier
\section{Active Learning Baselines} 
\label{app:active_learning}
We compare our method to typical methods used in the Active Learning (AL) literature. Note that our method is label-aware, while active learning acquires datapoints without using label information. We consider the following baselines, which select the top-$k$ points using an acquisiton function, $\alpha(x)$:

\begin{itemize}
    \item Bayesian Active Learning by Disagreement \citep{houlsby2011bayesian} with $\alpha(x) = \Hof{y \given x, \mathcal{D}_t} - \E{\pof{\theta \given \mathcal{D}_t}}{ \Hof{y \given x, \theta}}$.
    \item (Average) conditional entropy, $\alpha(x) = \E{\pof{\theta \given \mathcal{D}_t}}{ \Hof{y \given x, \theta}}$, where the average is taken over the model parameter posterior. 
    \item (Average predictive) entropy, $\alpha(x) = \Hof{y \given x, \mathcal{D}_t}$.
    \item Loss minus conditional entropy $\alpha(x) = \Lof{y \given x, \theta} - \E{\pof{\theta \given \mathcal{D}_t}}{ \Hof{y \given x, \theta}}$. This uses the (average) conditional entropy as an estimate of how noisy datapoint $x$ is---points with high noise are deprioritized. Compared to RHO-LOSS, it replaces the IL with the conditional entropy. This acquisition function uses the label and therefore cannot be used for active learning.
\end{itemize}
We additionally compare our method to uniform sampling. We run all baselines on MNIST and CIFAR10. Note that several of these active learning baselines consider epistemic uncertainty; that is, uncertainty in predictions driven by uncertainty in the model parameters. This mandates performing (approximate) Bayesian inference. We use Monte-Carlo Dropout\citep{gal2016dropout} to perform approximate inference. For MNIST, we use an 2 hidden layer MLP with 512 hidden units per hidden layer, and a dropout probability of a 0.5. For experiments on CIFAR10, we use a small-scale CNN with dropout probability 0.05 (the dropout probability follows \citep{osawa2019practical}).

\begin{figure}[h]
    \centering
    \includegraphics[width=\linewidth]{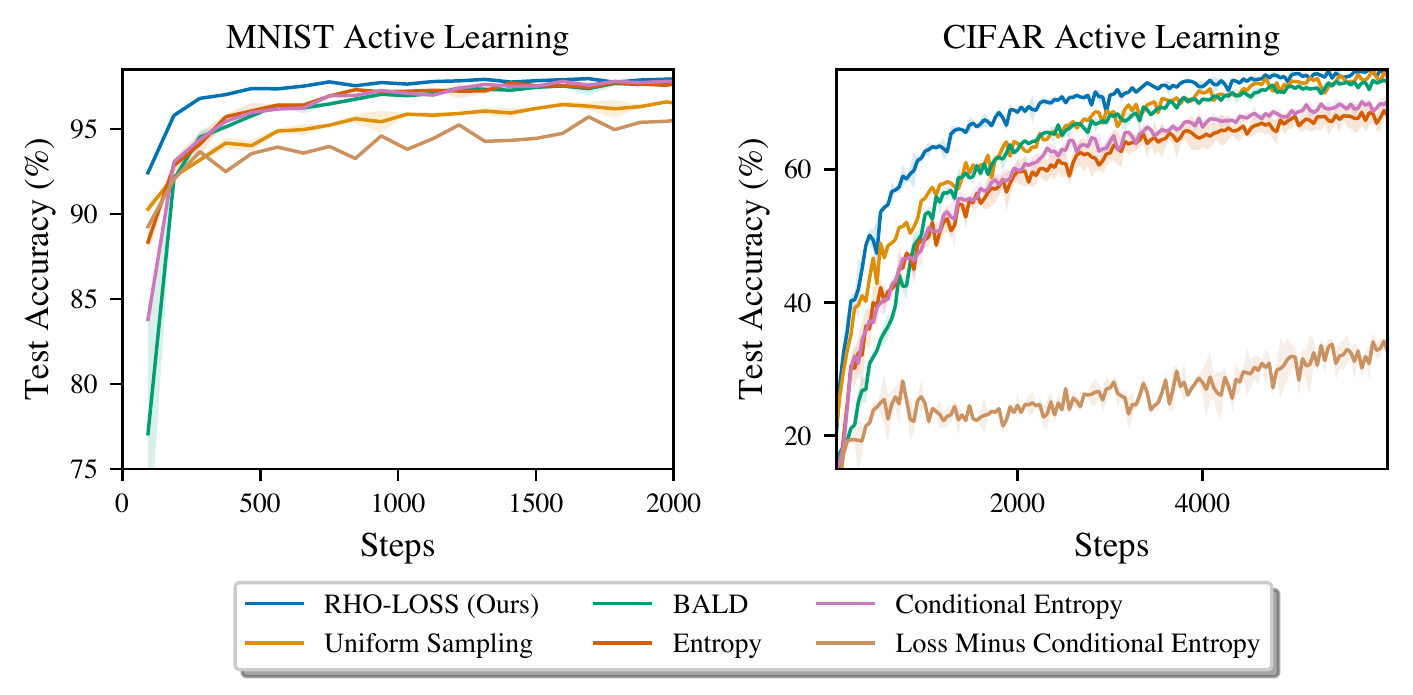}
    \caption{Training curves for several active learning baselines on the MNIST and CIFAR10 datasets.}
    \label{fig:active_learning_baselines}
\end{figure}

Fig.~\ref{fig:active_learning_baselines} shows training curves for our method, uniform sampling, and the active learning baselines. Our method accelerates training across both datasets. The active learning methods accelerate training for MNIST but not for CIFAR10. This highlights that active learning methods, if naively applied to online batch selection, may not accelerate model training.   

\end{document}